\pdfoutput=1
\documentclass[11pt]{article}

\usepackage[preprint]{acl}

\usepackage{times}
\usepackage{latexsym}

\usepackage[T1]{fontenc}
\usepackage[utf8]{inputenc}

\usepackage{microtype}

\usepackage{inconsolata}

\usepackage{graphicx}

\usepackage{amsmath}
\usepackage{amssymb}
\usepackage{booktabs}
\usepackage{xcolor}
\usepackage{pifont}
\usepackage{float}
\usepackage{subcaption}
\usepackage{booktabs}
\usepackage{multirow}
\usepackage{arydshln}
\usepackage{fontawesome5}
\definecolor{Green}{RGB}{0, 100, 0}
\definecolor{Bittersweet}{RGB}{220, 70, 20} 
\definecolor{Blue}{RGB}{0, 0, 139} 

\newcommand{\cmark}{\textcolor{Green}{\ding{51}}} 
\newcommand{\xmark}{\textcolor{Bittersweet}{\ding{55}}}
\newcommand{\bmark}{\textcolor{Blue}{\ding{72}}} % Star in blue

\newcommand{\pa}{DM-Codec}

\newcommand{\newtext}{\textcolor{black}} 
\newcommand{\acltext}{\textcolor{black}}

\title{DM-Codec: \textbf{\underline{D}}istilling \textbf{\underline{M}}ultimodal Representations for Speech Tokenization}

\author{\textbf{Md Mubtasim Ahasan\textsuperscript{1}\thanks{Corresponding author: mubtasimahasan@gmail.com}, 
Md Fahim\textsuperscript{1}, 
Tasnim Mohiuddin\textsuperscript{3},} \\
\textbf{A K M Mahbubur Rahman\textsuperscript{1}, 
Aman Chadha\textsuperscript{2}\thanks{Work does not relate to position at Amazon.}, 
Tariq Iqbal\textsuperscript{4},} \\
\textbf{M Ashraful Amin\textsuperscript{1}, 
Md Mofijul Islam\textsuperscript{2,4$\dagger$}\thanks{Equal Supervision.}, 
Amin Ahsan Ali\textsuperscript{1$\ddagger$}}\\
\textsuperscript{1}Center for Computational \& Data Sciences, Independent University, Bangladesh \\
\textsuperscript{2}Amazon GenAI
\textsuperscript{3}Qatar Computing Research Institute
\textsuperscript{4}University of Virginia \\
\small{\href{https://github.com/mubtasimahasan/DM-Codec}{\faGithub\enspace github.com/mubtasimahasan/DM-Codec}}
}
\begin{document}
\maketitle
\begin{abstract}

Recent advancements in speech-language models have yielded significant improvements in speech tokenization and synthesis. However, effectively mapping the complex, multidimensional attributes of speech into discrete tokens remains challenging. This process demands acoustic, semantic, and contextual information for precise speech representations. Existing speech representations generally fall into two categories: acoustic tokens from audio codecs and semantic tokens from speech self-supervised learning models. Although recent efforts have unified acoustic and semantic tokens for improved performance, they overlook the crucial role of contextual representation in comprehensive speech modeling. Our empirical investigations reveal that the absence of contextual representations results in elevated Word Error Rate (WER) and Word Information Lost (WIL) scores in speech transcriptions. To address these limitations, we propose two novel distillation approaches: (1) a language model (LM)-guided distillation method that incorporates contextual information, and (2) a combined LM and self-supervised speech model (SM)-guided distillation technique that effectively distills multimodal representations (acoustic, semantic, and contextual) into a comprehensive speech tokenizer, termed DM-Codec. The DM-Codec architecture adopts a streamlined encoder-decoder framework with a Residual Vector Quantizer (RVQ) and incorporates the LM and SM during the training process. Experiments show DM-Codec significantly outperforms state-of-the-art speech tokenization models, reducing WER by up to 13.46\%, WIL by 9.82\%, and improving speech quality by 5.84\% and intelligibility by 1.85\% on the LibriSpeech benchmark dataset. 
Code, samples, and checkpoints are available at 
\href{https://github.com/mubtasimahasan/DM-Codec}{\texttt{github.com/mubtasimahasan/DM-Codec}}

\end{abstract}

% Sections:

\section{Introduction}
\label{sec:intro}

% \looseness=-1
In recent years, the advent of Large Language Models (LLMs) has revolutionized various domains, offering unprecedented advancements across a wide array of tasks \citep{openai2024gpt4technicalreport}. A critical component of this success has been the tokenization of input data, enabling vast amounts of information processing \citep{discretetoken, goodtokenizer}. Inspired by these breakthroughs, significant attention has shifted towards replicating similar successes in the realm of speech understanding and generation \citep{encodec, hubert}. However, tokenizing speech into discrete units presents unique challenges compared to text, as speech is inherently continuous and multidimensional, requiring various speech attributes such as acoustic properties, semantic meaning, and contextual clues \citep{naturalspeech3}. Traditional approaches using feature representations such as Mel-Spectrograms \citep{highqualitysynthesis}, Mel-frequency cepstral coefficients (MFCCs) \citep{mfccsynthesis}, and Waveforms \citep{waveformsynthesis} have proven inadequate in capturing this full spectrum of information, resulting in suboptimal performance in downstream tasks such as speech synthesis \citep{naturalspeech3}. 

%%%%%%%%%%%%%%%%%% Figure %%%%%%%%%%%%%%%%%%%%%%%
\begin{figure*}[!t]
\begin{center}
    \includegraphics[width=0.85\linewidth]{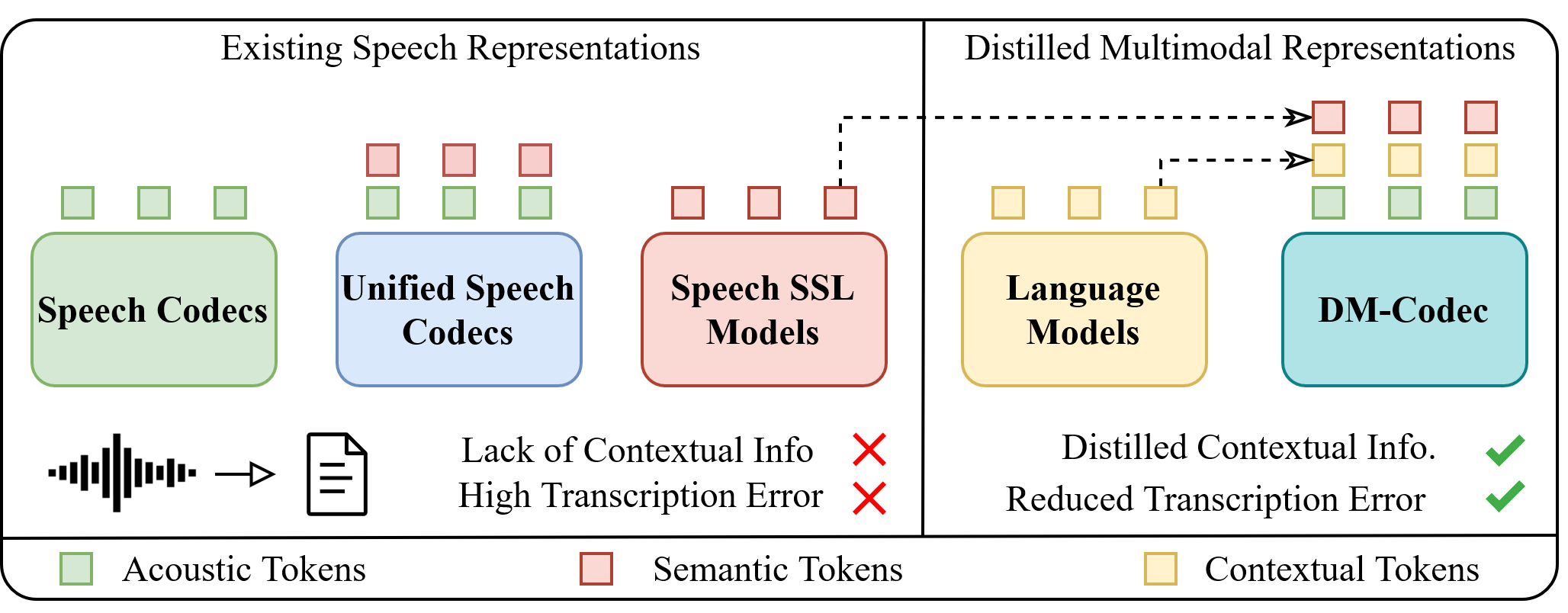} 
\end{center}
\caption{Overview of speech tokenization using discrete acoustic, semantic, and contextual tokens. DM-Codec integrates these features for robust and comprehensive speech representation.}
\label{fig:introfig}
% \vspace{-0.3cm}
\end{figure*}
%%%%%%%%%%%%%%%%%%%%%%%%%%%%%%%%%%%%%%%%%%%%%%%%%%

% \looseness=-1
These limitations led researchers to explore various approaches, and one prominent direction leading to audio codecs \citep{borsos2023audiolm}. Notable examples include SoundStream \citep{soundstream} and EnCodec \citep{encodec}, which utilize Residual Vector Quantizers (RVQ) within a neural codec framework, iteratively refining quantized vectors to discretize speech into \textit{acoustic} tokens. Concurrently, self-supervised speech representation learning models such as HuBERT \citep{hubert} and wav2vec 2.0 \citep{wav2vec20} facilitated extracting speech representations as \textit{semantic} tokens \citep{borsos2023audiolm}. Efforts to unify acoustic and semantic representations have led to two notable approaches: SpeechTokenizer \citep{speechtokenizer}, which utilizes semantic distillation from HuBERT, and FACodec \citep{naturalspeech3}, which proposes a factorized vector quantizer to disentangle speech representation into different subspaces using separate RVQs with supervision. 

% % \looseness=-1
However, these approaches often overlook a key aspect: the integration of \textit{contextual} information. Language models (LMs) have demonstrated an ability to learn contextual representations that capture the meaning of tokens from broader context \citep{bert}. These contextual representations can provide essential insights into speech representation, allowing for a more nuanced understanding of words in varying linguistic contexts. Our empirical investigations also reveal that existing discrete speech representation models struggle to align reconstructed speech with accurate textual form, resulting in elevated Word Error Rates (WER) and Word Information Lost (WIL) scores in speech transcription. This observation underscores the need for a more comprehensive approach to speech tokenization that incorporates contextual information.

% \looseness=-1
To address these challenges, we propose {\pa}, a novel speech tokenizer that unifies multimodal language and speech representations. 
Central to our innovation is the introduction of an LM-guided distillation method that incorporates contextual representations into the speech tokenization process. This allows {\pa} capturing the nuances of linguistic context often missed by existing models. Building upon the LM-guided approach, we propose a distillation method combining both LM and speech model (SM)-guided techniques. \newtext{Moreover, we introduce a [CLS]-token-based distillation strategy that leverages sequence-level holistic representations from the LM, effectively capturing global contextual information.
Our distillation method only utilizes the LM and SM during training, without increasing model complexity and parameters, and are not required during inference.
} To the best of our knowledge, we are the first to attempt to integrate all three essential aspects of speech representation—acoustic, semantic, and contextual—within a single codec. See Figure \ref{fig:introfig} for a depiction. \newtext{In addition, to demonstrate the impact of multimodal representation and generalizability of {\pa} in downstream tasks, we introduce {\pa}-TTS, a novel multimodal representation distilled neural codec language model.}

% \looseness=-1
Through extensive experiments on LibriSpeech \citep{librispeech}, we show DM-Codec’s superiority. DM-Codec achieves a WER of $4.05$ and WIL of $6.61$, outperforming SpeechTokenizer ($4.49$, $7.10$), FACodec ($4.68$, $7.33$), and EnCodec ($4.53$, $7.17$). It also improves speech quality, with a ViSQOL score of $3.26$ and MOS of $3.72$, surpassing EnCodec ($3.08$, $3.09$), SpeechTokenizer ($3.09$, $3.67$), and FACodec ($3.13$, $3.70$). DM-Codec-TTS excels on LibriSpeech and VCTK, outperforming USLM and Vall-E. On LibriSpeech, it achieves a WER of $5.08$, WIL of $7.32$, MOS of $3.70$, and SMOS of $3.89$, while on VCTK, it achieves a WER of $3.58$, WIL of $5.65$, MOS of $3.78$, and SMOS of $3.85$. Notably, DM-Codec-TTS-small achieves a WER of $10.26$, WIL of $13.79$, MOS of $3.24$, and SMOS of $3.20$, outperforming USLM (libri) across all metrics despite using a smaller dataset.

We summarize our contributions below:

\begin{itemize}
    \item We introduce {\pa}, a novel speech tokenizer that incorporates contextual representations via the LM-guided distillation method.
    \item We present a novel combined LM and SM-guided distillation method, unifying acoustic, semantic, and contextual representations.
    \item \newtext{We propose a [CLS]-token-based distillation strategy that captures global contextual information from the LM, facilitating better alignment and transfer of contextual features.}
    \item \newtext{We introduce DM-Codec-TTS, a neural codec language model, demonstrating the generalizability of the DM-Codec framework in downstream tasks.}
\end{itemize}

\section{Related Work} 
\label{sec:related_work}

\textbf{Tokenization Techniques in Speech.} Tokenization in speech processing can be broadly categorized into two main approaches: (i) speech encoder-based and (ii) language-based. In the speech encoder-based tokenization approach, a pretrained speech encoder serves as a teacher model, providing semantically rich audio representations. These representations are then used to guide the training model, either through an alignment network \citep{messica2024nast} or by optimizing specific losses \citep{speechtokenizer,liu2024dinosr}. Language-based tokenization approach involves processing audio through a speech encoder to obtain discrete representations or using the corresponding text to feed into a language model. The representations from the language model are then utilized either to learn a tokenizer for speech or to reconstruct speech \citep{last2024, hassid2024textually, zhang2024speechlm, wang2024selm}.  Besides, \citep{zhang2024speechlm} proposed SpeechLM where two discrete tokenizers were introduced and learned in an unsupervised way and converted the speech and text in a shared discrete space.

\textbf{Discrete Speech Representation.} There are two well-known methods for discrete speech representation: semantic tokens and acoustic tokens. Semantic tokens are derived through self-supervised learning (SSL) techniques for speech \citep{baevski2019vq, hubert, chung2021w2v} and capture abstract, high-level features that relate to general, symbolic aspects of speech, while omitting details related to speaker identity and acoustic characteristics. In contrast, acoustic tokens are obtained using neural audio codecs \citep{soundstream, encodec, hificodec} and focus on delivering precise reconstructions of acoustic features. However, recent models \citep{last2024, liu2024dinosr, shi2024mmm} have shown that speech models based on self-supervised learning (SSL) are effective at extracting acoustic representations where LMs be employed to refine these models further, enhancing their ability to extract more nuanced semantic representations. \acltext{Recent works \citep{repcodec, cosyvoice, fusecodec} focused on improving the quality of discrete speech representation by retaining semantic information for both speech understanding and generation.}

\textbf{Textual Language Models in Speech.}  Research on speech models, including works by \citep{nguyen2023generative}, \citep{borsos2023audiolm}, and \citep{kharitonov-etal-2022-text}, has focused on utilizing raw audio to extract prosodic features, identify speaker characteristics, and generate audio without depending on textual features or supervision from textual LMs. In contrast, many newer methods have started using audio encoders to transform audio signals into discrete tokens, which can be processed by textual LMs. TWIST method introduced by \citep{hassid2024textually} initializes the weights of the SpeechLM using a pre-trained text LM, showing that this combination significantly improves performance. Similarly, the SELM model developed by \citep{wang2024selm} leverages GPT \citep{radford2018improving, radford2019language} as its foundation due to its enhanced parallel processing capabilities and capacity. However, text-based LLMs such as GPT-3 \citep{brown2020language} and Llama \citep{touvron2023llama} are essential for speech modeling. Once discrete audio representations are obtained, these large text models are trained to enhance the original text embedding space, as explored in studies by \citep{zhang2023speechgpt}, \citep{fathullah2023towards}, \citep{shu2023llasm}, and \citep{rubenstein2023audiopalm}. This trend of integrating textual LMs into speech modeling has become increasingly popular in recent research.

%%%%%%%%%%%%%%%%%%%%%%%%%%%%%%%%%%%%%%%%%%%%%%%%%%%%%%%%%%%
\begin{figure*}[!t]
\begin{center}
    \includegraphics[width=0.9\linewidth]{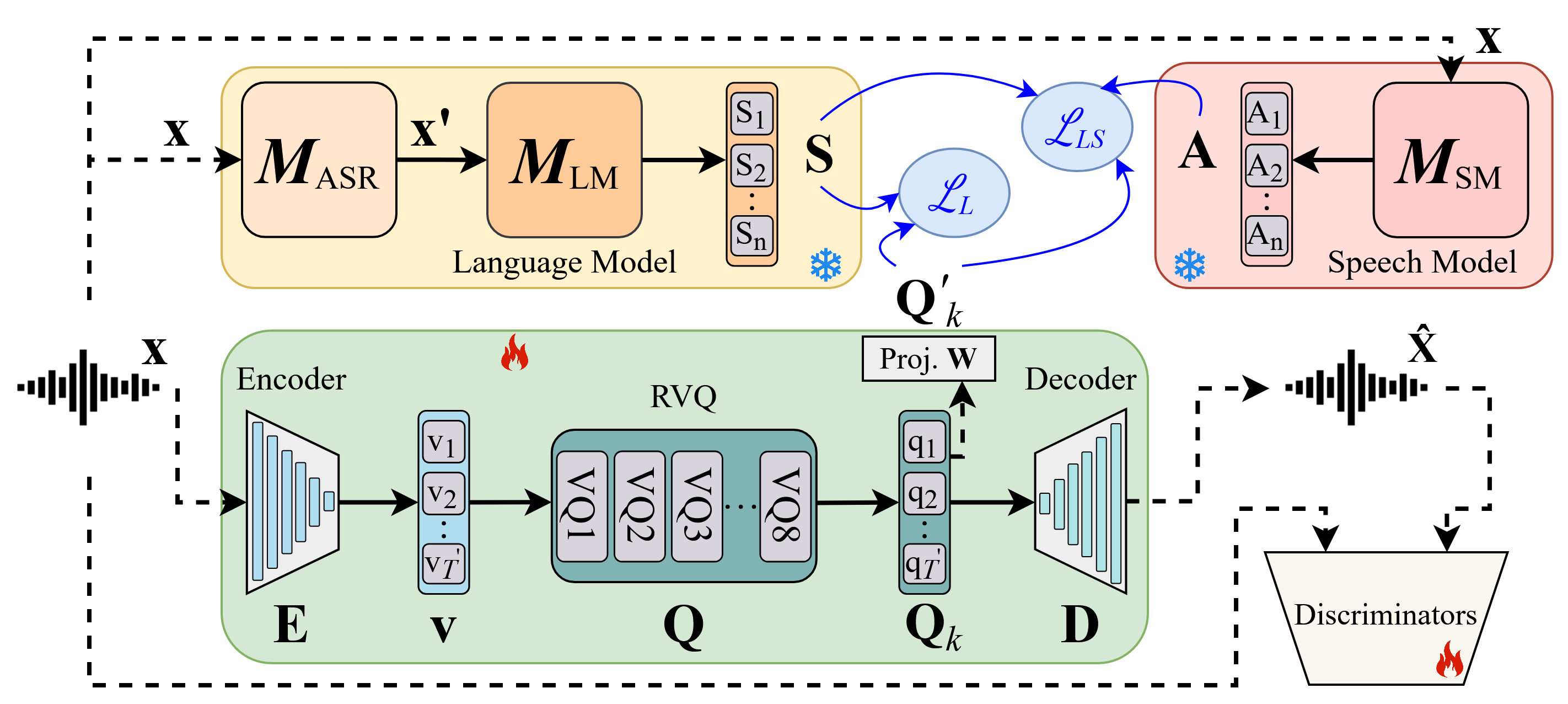}
\end{center}
\caption{{\pa} framework consists of an encoder that extracts latent speech representations, and quantized using a Residual Vector Quantizer (RVQ). We propose two distillation approaches: (i) from a language model (LM), and (ii) from both an LM and a speech model (SM), integrating acoustic, semantic, and contextual features to enhance speech representations for downstream tasks.}
\label{fig:tokenizer}
% % \vspace{-10pt}
\end{figure*}
%%%%%%%%%%%%%%%%%%%%%%%%%%%%%%%%%%%%%%%%%%%%%%%%%%%%%%%%%%%

\section{Proposed Method}
\label{sec:method}
% % \vspace{-5pt}
 
% \looseness=-1
This section introduces {\pa}, a novel speech tokenizer that distills multimodal (acoustic, semantic, and contextual) representations. As shown in Figure \ref{fig:tokenizer}, we propose two training approaches: (i) LM-guided distillation, integrating contextual and acoustic representations, and (ii) combined LM and SM-guided distillation, incorporating semantic, contextual, and acoustic representations. Additionally, a [CLS] token-guided distillation leverages sequence-level holistic contextual representations. We also present DM-Codec-TTS, integrating DM-Codec into a neural codec language model. The following subsections detail the distillation methods (\S\ref{sec:semantic}), DM-Codec model and training objectives (\S\ref{sec:details}, \S\ref{sec:train_obj}), and DM-Codec-TTS (\S\ref{sec:dmcodec_tts}).

\subsection{Multimodal Representation Distillation}
\label{sec:semantic}
\looseness=-1
We first transcribe the raw speech \( \mathbf{x} \) into its corresponding text \( \mathbf{x'} \) using an \newtext{Automatic Speech Recognition (ASR) model \( M_{ASR} \), such that \(\mathbf{x'}\,=\,M_{ASR}(\mathbf{x}) \)}. \acltext{\( M_{ASR} \) serves as an tool for converting speech to text, and provides no supervision or influence on model training}. For simplicity, we omit any post-processing techniques on the \( \mathbf{x'} \). Subsequently, we pass the text \( \mathbf{x'} \) through a pretrained language model \( M_{LM} \) to obtain contextual representations of \( \mathbf{x'} \), tokenized into a set of tokens, \( \mathcal{T} = \{t_i\}_{i=1}^n \). For each token \( t_i \), we extract its corresponding layer-wise hidden representations \( \{\mathbf{h}_i^l\}_{l=1}^L \), where \( L \) denotes the total number of layers in \( M_{LM} \). We utilize all layer representations to derive the representations for each token, as each layer of a pre-trained language model captures hierarchical and contextually distinct information \citep{niu2022does, kovaleva2019revealing, hao2019visualizing}. To obtain the contextual representation \( \mathbf{S}_i \) for token \( t_i \), we average the hidden representations across all layers, yielding $ \mathbf{S}_i = \frac{1}{L} \sum_{l=1}^{L} \mathbf{h}_i^l$, where \( \mathbf{S}_i \in \mathbb{R}^{D} \) where $D$ is the hidden dimension. Consequently, we obtain the contextual representations \( \mathbf{S} = \{\mathbf{S}_i\}_{i=1}^n \) for the speech input \( \mathbf{x} \), which captures the contextually diverse information from \( M_{LM} \).

\looseness=-1
Simultaneously, we process the raw speech \( \mathbf{x} \) through an Encoder \( \mathbf{E}(\mathbf{x}) \) to obtain the latent features \( \mathbf{v} \), \newtext{with sequence length \( T' \)}. We then pass \( \mathbf{v} \) through a Residual Vector Quantizer (RVQ) to obtain quantized features \( \mathbf{Q} = \{\mathbf{Q}_k\}_{k=1}^K \), where \( K \) represents the number of quantization layers in the RVQ, and \( \mathbf{Q}_k \in \mathbb{R}^{D'} \) where $D'$ is the hidden dimension of $k^{th}$ RVQ layer. 
\acltext{\( \mathbf{Q}_k\) capture comprehensive speech information across the entire utterance and are subsequently used to reconstruct the audio \( \mathbf{\hat{x}} \) via the decoder. The holistic information of the entire speech segment, represented in discrete form within \( \mathbf{Q}_k \), allows us to directly match it with \( \mathbf{S} \) by padding the tokens \( n \) to the quantized sequence length \( T' \). The padded tokens serve only to measure feature dimension similarity rather than impose temporal alignment. To distill contextual information from \( \mathbf{S} \) into \( \mathbf{Q}_k \), we apply a linear transformation $\mathbf{Q}_k' = \mathbf{W} \mathbf{Q}_k$, where \( \mathbf{W} \in \mathbb{R}^{D' \times D} \), ensuring the dimensional consistency for alignment and updates to occur within each feature dimension without necessitating strict temporal correspondence.}

\textbf{LM Guided Distillation:} 
\looseness=-1
In this approach, we distil the LM representations $\mathbf{S}$. \acltext{To calculate the distillation loss, we adopt \textit{continuous representation distillation} \citep{speechtokenizer}, which maximizes the cosine similarity at the feature dimension axis $D$ with the motivation to focus on similarity within each feature dimension $D$ rather than solely focusing on the overall output similarity.} We calculate the continuous representation distillation of the transformed quantized features \( \mathbf{Q}_k' \) and the LM representation features \( \mathbf{S} \) as follows:

% \vspace{-0.2cm}
\begin{equation}
    \mathcal{L}_{L} = - \frac{1}{D} \sum_{d=1}^{D} \log\left(\sigma\left(\frac{\mathbf{Q}_k'^{(:,d)} \cdot \mathbf{S}^{(:,d)}}{\|\mathbf{Q}_k'^{(:,d)}\|  \|\mathbf{S}^{(:,d)}\|}\right)\right)
    \label{eq:lm_distil} 
\end{equation}
% \vspace{-0.2cm}

\looseness=-1
Here, the notation \((:, d)\) indicates a vector that includes values from all time steps at the \( d ^{th} \) dimension, \acltext{where \( d \in [1, D] \) in the \( D \)-dimensional feature space. The function \(\sigma(\cdot)\) represents the sigmoid activation function. \(\sum_{d=1}^{D}\) computes the loss by aggregating over all feature dimensions, ensuring that updates occur at the feature level rather than at specific time steps.}

\textbf{Combined LM and SM Guided Distillation:} 
\looseness=-1
We enhance {\pa} with a hybrid approach using both audio and text modalities. To derive semantic representations from the speech model (SM), we adopt a similar distillation strategy as we used for the LM. We first pass the raw speech $\mathbf{x}$ through the pretrained speech model \( M_{SM} \), generating layer-wise hidden representations \( \{\mathbf{h}_j^l\}_{l=1}^L \). The semantic features are derived by averaging the hidden states across all layers, yielding $
\mathbf{A}_j = \frac{1}{L} \sum_{l=1}^{L} \mathbf{h}_j^l$, where \( \mathbf{A}_j \in \mathbb{R}^{D} \). This yields semantic representations \( \mathbf{A} = \{\mathbf{A}_j\}_{j=1}^n \) for speech input \( \mathbf{x} \). The distillation loss in this case considers both the LM and SM representations, jointly optimizing the quantized features  \( \mathbf{Q}_k' \) with the representations $\mathbf{A}$ and $\mathbf{S}$ derived from $M_{SM}$ and $M_{LM}$, respectively. We first calculate the distillation loss for the SM, \( \mathcal{L}_{S} \), followed by averaging with the LM distillation loss, \( \mathcal{L}_{L} \), to ensure a balanced contribution from both losses, as follows: 

% \vspace{-0.2cm}
\begin{equation}
    \mathcal{L}_{S} = -\frac{1}{D} \sum_{d=1}^{D} \log\left(\sigma\left(\frac{\mathbf{Q}_k'^{(:,d)} \cdot \mathbf{A}^{(:,d)}}{\|\mathbf{Q}_k'^{(:,d)}\| \|\mathbf{A}^{(:,d)}\|}\right)\right)
\label{eq:distil} 
\end{equation}

\begin{equation}
    \mathcal{L}_{LS} = \frac{1}{2} \left(\mathcal{L}_{L} + \mathcal{L}_{S} \right)
\label{eq:combined_distil} 
\end{equation}
% \vspace{-0.2cm}

This formulation ensures that {\pa} integrates acoustic representation learned by the codec architecture with semantic knowledge from SM and contextual knowledge from LM.

\newtext{\textbf{[CLS] Token Guided Distillation:}}
\newtext{We introduce a [CLS] token-guided distillation strategy, leveraging the [CLS] token's sequence-level holistic representation to capture global contextual information from LM. This approach eliminates the need for fine-grained temporal alignment while preserving essential linguistic features.} Moreover, CLS-guided distillation employs sentence-level knowledge transfer, which has been shown to be more robust in noisy settings \citep{cls}. \newtext{For this method, we use the layer-wise hidden representations of the [CLS] token alone. These representations, averaged across all layers, are denoted as \( \mathbf{S}_{\text{[CLS]}} = \frac{1}{L} \sum_{l=1}^{L} \mathbf{h}_{\text{[CLS]}}^l \), where \( \mathbf{S}_{\text{[CLS]}} \in \mathbb{R}^{D} \). To match the sequence length \( T' \) of the quantized features \( \mathbf{Q}_k' \), the [CLS] token representation is repeated \( T' \) times, yielding \( \mathbf{S}' = \{\mathbf{S}_{\text{[CLS]}}, \mathbf{S}_{\text{[CLS]}}, \ldots, \mathbf{S}_{\text{[CLS]}} \} \).} 

\newtext{The distillation loss follows the same formulation as the LM-guided distillation loss (Eqn.~\ref{eq:lm_distil}), replacing \( \mathbf{S} \) with \( \mathbf{S}' \). This enables the distillation process to leverage the global contextual information encoded in the [CLS] token while ensuring greater alignment with the sequence length \( T' \).}

% \vspace{-2pt}
\subsection{\newtext{DM-Codec:} Model Details}
\label{sec:details}
% \vspace{-2pt}
Our framework builds on the Residual Vector Quantizer with Generative Adversarial Networks (RVQ-GAN) architecture. The core model comprises an Encoder \( \mathbf{E} \) and Decoder \( \mathbf{D} \) with an RVQ structure, inspired by Encodec \citep{encodec} and SpeechTokenizer \citep{speechtokenizer}. \acltext{We utilize a codebook size of 1024 and 8 quantization levels at a 50Hz frame rate.} We employ a multi-discriminator setup, including Multi-Scale Discriminator (MSD), Multi-Period Discriminator (MPD), and Multi-Scale Short-Time Fourier Transform (MS-STFT) Discriminator, drawn from HiFi-Codec \citep{hificodec} and HiFi-GAN \citep{hifigan}. Detailed architectural specifications are in Appendix \ref{sec:components}. To enhance the quantizer with distilled multimodal representations, we use wav2vec 2.0 (wav2vec2-base-960h) as \( M_{ASR} \) \citep{wav2vec20}, BERT (bert-base-uncased) as \( M_{LM} \) \citep{bert}, and HuBERT (hubert-base-ls960) as \( M_{SM} \) \citep{hubert}. Quantized outputs from the first RVQ layer (RVQ-1) are used for LM-guided distillation, and the average of quantized outputs across all eight layers (RVQ-1:8) is used for SM-guided distillation. An ablation study of RVQ layers selection is in Appendix \ref{sec:layers}.

% % \vspace{-1pt}
\subsection{\newtext{DM-Codec:} Training Objective}
\label{sec:train_obj}
% % \vspace{-1pt}

In addition to the distillation losses in Section \ref{sec:semantic}, our training strategy builds on methodologies \citep{speechtokenizer, hificodec}, employing the RVQ-GAN framework. For the original speech \( \mathbf{x} \) and the reconstructed speech \( \hat{\mathbf{x}} \), we use reconstruction, adversarial, feature matching, and commitment losses to guide learning as follows:

\textbf{Reconstruction Loss.} To ensure that the model preserves the key attributes of speech, we employ both time-domain and frequency-domain reconstruction losses. The time-domain loss \( \mathcal{L}_t \) is computed as the L1 distance between \( \mathbf{x} \) and \( \hat{\mathbf{x}} \). For the frequency-domain loss \( \mathcal{L}_f \), we combine L1 and L2 losses over 64-bin Mel-spectrograms \( \text{Mel}_i \), with varying window sizes of \( 2^i \),  hop lengths of \( 2^i / 4 \), and scales \( e = \{5, \ldots, 11\} \).

% \vspace{-0.3cm}
\begin{equation}
    \mathcal{L}_{\text{t}} = \| \mathbf{x} - \hat{\mathbf{x}} \|_1
\end{equation}

\begin{equation}
\begin{aligned}
    \mathcal{L}_f &= \sum_{i \in e} \Big( 
    \| \text{Mel}_i(\mathbf{x}) - \text{Mel}_i(\hat{\mathbf{x}}) \|_1 \\
    &\quad+ \| \text{Mel}_i(\mathbf{x}) - \text{Mel}_i(\hat{\mathbf{x}}) \|_2 
    \Big)
\end{aligned}
\end{equation}

% % \vspace{-0.3cm}

\textbf{Adversarial Loss.} The adversarial loss encourages the generator to produce realistic, indistinguishable speech. We apply a hinge loss formulation to compute the adversarial loss for the generator \( \mathcal{L}_g \) and the discriminator \( \mathcal{L}_d \). These losses are computed across all three discriminators: the multi-scale discriminator (MSD), multi-period discriminator (MPD), and the multi-scale STFT guided (MS-STFT) discriminator (see Appendix \ref{sec:components}).

% \vspace{-0.3cm}

\begin{equation}
    \mathcal{L}_g = \frac{1}{N} \sum_{n=1}^{N} \max(1 - R_n(\hat{\mathbf{x}}), 0)
\end{equation}

\begin{equation}
\begin{aligned}
    \mathcal{L}_d &= \frac{1}{N} \sum_{n=1}^{N} \Big( 
    \max(1 - R_n(\mathbf{x}), 0) \\
    &\quad+ \max(1 + R_n(\hat{\mathbf{x}}), 0) 
    \Big)
\end{aligned}
\end{equation}

% % \vspace{-0.3cm}

where \( N \)  is the number of discriminators and \( R_n \) represents the $n^{th}$ discriminator.

\textbf{Feature Matching Loss.} To prevent the generator from overfitting to the discriminator's decisions, we apply a feature matching loss \( \mathcal{L}_{fm} \). This loss compares features from each discriminator \( R_n \)’s internal layers \( M \) across all dimensions, promoting stability and better generalization.

% \vspace{-0.3cm}

\begin{equation}
\mathcal{L}_{fm} = \frac{1}{NM} \sum_{n=1}^{N} \sum_{m=1}^{M} \frac{\| R_{n}^{m}(\mathbf{x}) - R_{n}^{m}(\hat{\mathbf{x}}) \|_1}{\text{mean}(\| R_{n}^{m}(\mathbf{x}) \|_1)}    
\end{equation}

% % \vspace{-0.3cm}

\textbf{RVQ Commitment Loss.} To guide the encoder to produce outputs that closely match their corresponding quantized values in the residual vector quantization (RVQ) process, we introduce a commitment loss \( \mathcal{L}_w \). For \( N_q \) quantization vectors, where \( \mathbf{q}_i \) represents the current residual and \( \mathbf{q}_{c_i} \) is the closest entry in the corresponding codebook for the \( i ^{th}\) entry, the \( \mathcal{L}_w \) is computed as:

% \vspace{-0.3cm}
\begin{equation}
\mathcal{L}_w = \sum_{i=1}^{N_q} \| \mathbf{q}_i - \mathbf{q}_{c_i} \|_2^2
\end{equation}

\textbf{Overall Generator Loss.} The total generator loss \( \mathcal{L}_G \) is a weighted sum of the individual loss components, including the distillation loss \(\mathcal{L}_{L/LS} \) (which is either \(\mathcal{L}_{L} \) or \(\mathcal{L}_{LS} \) depending on the chosen distillation method). We use the corresponding weighting factors \(\lambda_{L/LS}, \lambda_t, \lambda_f, \lambda_g, \lambda_{fm}\), and \( \lambda_w \) to control the influence of each loss component on the overall training objective as: 

% \vspace{-0.3cm}

\begin{equation}
\begin{aligned}
    \mathcal{L}_G &= \lambda_{L/LS} \mathcal{L}_{L/LS} 
    + \lambda_t \mathcal{L}_t 
    + \lambda_f \mathcal{L}_f \\
    &\quad+ \lambda_g \mathcal{L}_g 
    + \lambda_{fm} \mathcal{L}_{fm} 
    + \lambda_w \mathcal{L}_w
\end{aligned}    
\end{equation}

This comprehensive training objective ensures {\pa} learns acoustic speech representations while incorporating semantic and contextual representation through novel distillation approaches.

\subsection{\newtext{DM-Codec-TTS: Model Details and Training Objective}}
\label{sec:dmcodec_tts}
% % \vspace{-1pt}
\newtext{Following SpeechTokenizer \citep{speechtokenizer} and VALL-E \citep{valle}, we propose DM-Codec-TTS, a novel multimodal representation distilled neural codec-based Text-To-Speech (TTS) model. Extending upon general neural codec language models, DM-Codec-TTS, leverages the strength of contextual and semantic representation distilled on our neural codec, DM-Codec.}

\newtext{\textbf{Problem Formulation.}
For zero-shot TTS, the task is to synthesize speech for a given speaker. We frame it as a conditional codec language modeling problem, where the objective of DM-Codec-TTS is to predict the quantized acoustic features \( \mathbf{Q} = \{\mathbf{Q}_k\}_{k=1}^K \), conditioned on a phoneme sequence \( \mathbf{u} \) and an acoustic prompt \( \tilde{\mathbf{P}} \in \mathbb{R}^{T' \times K} \) extracted from the input enrolled recording. }

\newtext{\textbf{Training Objective.}
The model integrates autoregressive (AR) and non-autoregressive (NAR) components to hierarchically encode speech information. The AR component models content and speaker identity by predicting tokens \( \mathbf{Q}_1^t \) from the first RVQ layer using a transformer decoder-only architecture \( \phi_{\text{AR}} \). The AR training objective is:}

% \vspace{-0.3cm}

\begin{equation}
\mathcal{J}_{\text{AR}} = - \sum_{t=0}^{T} \log \, p\left(\mathbf{Q}_1^t \mid \mathbf{Q}_1^{<t}, \mathbf{u}; \phi_{\text{AR}}\right)    
\end{equation}

% % \vspace{-0.3cm}

\looseness=-1
\newtext{The NAR component focuses on acoustic details by predicting tokens \( \mathbf{Q}_k \) (\( k = 2, \dots, 8 \)) from subsequent RVQ layers. The NAR training objective is:}

% \vspace{-0.3cm}
\begin{equation}
\mathcal{J}_{\text{NAR}} = - \sum_{k=2}^{8} \log \, p\left(\mathbf{Q}_k \mid \mathbf{Q}_{<k}, \tilde{\mathbf{P}}, \mathbf{u}; \phi_{\text{NAR}}\right)
\end{equation}
% % \vspace{-0.3cm}

\newtext{During inference, the AR model predicts tokens \( \mathbf{Q}_1 \) based on speech phonemes \( \mathbf{u} \), while the NAR model iteratively generates \( \mathbf{Q}_{2:8} \) using the AR output and the acoustic prompt \( \tilde{\mathbf{P}} \). The combined tokens \( \mathbf{Q} = \{\mathbf{Q}_1, \mathbf{Q}_{2:8}\} \) are decoded into a speech waveform via the neural codec.}

\newtext{\textbf{Model Details.} The AR and NAR models share an identical transformer structure, comprising 12 layers, 16 attention heads, a 1024-dimensional embedding space, 4096-dimensional feed-forward layers, and a dropout rate of 0.1.}

%%%%%%%%%%%%%%%%%%%%%%%%%%%%%%%%%%%%%%%%%%%%%%%%%
%%%%%%%%%%%%%%%%%%%%%%%% START TABLE 1 %%%%%%%%%%%%%%%%%%%%%%%%%%%%%%%%%
\begin{table*}[hb!t]
\caption{Evaluation of DM-Codec's speech reconstruction quality against baselines. {\pa} (LM+SM) achieves the best performance, highlighting the impact of contextual and semantic representation distillation.\textsuperscript{$\heartsuit$} means the results were reproduced using the official training code. \textsuperscript{$\diamondsuit$} means the results were obtained using official model checkpoints.  (LM) indicates LM-guided distillation.  (LM+SM) indicates combined LM and SM-guided distillation. \newtext{(CLS) indicates [CLS]-token based distillation. Baseline indicates {\pa} without any distillation.} \acltext{(SM Baseline) indicates the SM only distillation baseline.} \textbf{Bold} highlights the best and \underline{underline} the second-best result.}
\label{table:reconstruction1}
\begin{center}
\resizebox{\textwidth}{!}{
% \small
\begin{tabular}{l cccccccc}
\toprule
\multicolumn{1}{c}{\bf Model} & \multicolumn{1}{c}{\bf WER $\downarrow$} & \multicolumn{1}{c}{\bf WIL $\downarrow$} & \multicolumn{1}{c}{\bf ViSQOL $\uparrow$} & \multicolumn{1}{c}{\bf STOI $\uparrow$} & \multicolumn{1}{c}{\bf Similarity $\uparrow$} & \multicolumn{1}{c}{\bf MOS $\uparrow$} & \multicolumn{1}{c}{\acltext{\bf UTMOS $\uparrow$}} & \multicolumn{1}{c}{\acltext{\bf PESQ $\uparrow$}} \\
\midrule
Groundtruth & 3.78 & 6.03 & - & - & - & - & \acltext{-} & \acltext{-} \\ 
EnCodec\textsuperscript{$\diamondsuit$} & 4.53 & 7.17 & 3.08 & 0.920 & 0.980 & 3.09 & \acltext{2.48} & \acltext{2.34} \\
SpeechTokenizer\textsuperscript{$\heartsuit$} & 4.49 & 7.10 & 3.09 & 0.923 & 0.993 & 3.67 & \acltext{3.49} & \acltext{2.63} \\ 
FACodec\textsuperscript{$\diamondsuit$} & 4.68 & 7.33 & 3.13 & \textbf{0.949} & \textbf{0.996} & 3.70 & \acltext{3.52} & \acltext{2.86} \\ 
\acltext{DAC\textsuperscript{$\diamondsuit$}} & \acltext{4.15} & \acltext{6.61} & \acltext{3.20} & \acltext{0.941} & \acltext{0.996} & \acltext{-} & \acltext{3.40} & \acltext{2.76} \\  
\acltext{BigCodec\textsuperscript{$\diamondsuit$}} & \acltext{4.54} & \acltext{7.43} & \acltext{3.02} & \acltext{0.937} & \acltext{0.996} & \acltext{-} & \acltext{3.51} & \acltext{2.71} \\  
\acltext{Mimi\textsuperscript{$\diamondsuit$}} & \acltext{11.78} & \acltext{18.34} & \acltext{2.49} & \acltext{0.853} & \acltext{0.936} & \acltext{-} & \acltext{2.35} & \acltext{1.70} \\  
% \hdashline
\midrule
\newtext{DM-Codec (Baseline)} & 4.97 & 8.02 & 2.95 & 0.935 & 0.991 & 3.13 & \acltext{3.25} & \acltext{2.58} \\  
\acltext{DM-Codec (SM Baseline)} & \acltext{4.49} & \acltext{7.25} & \acltext{3.12} & \acltext{0.933} & \acltext{0.994} & \acltext{-} & \acltext{3.43} & \acltext{2.86} \\
\newtext{DM-Codec (CLS)} & 4.47 & 7.08 & 3.12 & 0.926 & 0.993 & 3.65 & \acltext{3.39} & \acltext{2.65} \\ 
{\pa} (LM) & 4.36 & 7.06 & 3.18 & 0.935 & 0.994 & 3.69 & \acltext{3.48} & \acltext{\textbf{2.86}} \\  
\textbf{{\pa} (LM+SM)} & \textbf{4.05} & \textbf{6.61} & \textbf{3.26} & 0.937 & \underline{0.994} & \textbf{3.72} & \acltext{\textbf{3.52}} & \acltext{\underline{2.81}} \\ 
\bottomrule
\end{tabular}
}
\end{center}
% \vspace{-.5cm}
\end{table*}

%%%%%%%%%%%%%%%%%%%%%%%% End TABLE 1 %%%%%%%%%%%%%%%%%%%%%%%%%%%%%%%%%
%%%%%%%%%%%%%%%%%%%%%%%%%%%%%%%%%%%%%%%%%%%%%%%%%

\section{Experimental Setup}
% %\vspace{-5pt}
\textbf{Dataset.}
\looseness=-1
We trained {\pa} using the LibriSpeech 100-hour clean speech training set \citep{librispeech}, widely used for speech tokenizer and modeling \citep{speechtokenizer, naturalspeech3, hubert}. Data were standardized by cropping samples to three seconds and ensuring a 16 Hz sample rate. For DM-Codec-TTS, we used LibriHeavy \citep{libriheavy}, a 50k-hour read English speech dataset derived from LibriVox, selecting samples between 0.5 and 15 seconds at 16 kHz. Additionally, a smaller DM-Codec-TTS version was trained on LibriTTS \citep{libritts}, a 585-hour dataset with 2,456 speakers sampled at 24 kHz.  

For evaluation, we tested {\pa} on 300 randomly selected audio samples from the LibriSpeech test-clean subset, aligning with baseline Speechtokenizer's \citep{speechtokenizer} setup. A random seed of 42 ensured consistent sampling, and baseline models were evaluated on the same setup.  Additionally, we tested DM-Codec-TTS on LibriSpeech test-clean and VCTK \citep{vctk}. For LibriSpeech, we followed VALL-E's \citep{valle} setup by constructing a 2.2-hour subset from the test-clean data, selecting samples of 4–10 seconds. For each synthesis, a different utterance from the same speaker was randomly selected as input text, and a 3-second segment was cropped as the enrollment speech prompts. Each experiment was repeated three times, and the average score was reported. For VCTK, we followed SpeechTokenizer’s \citep{speechtokenizer} setup, using a 3s utterance as the prompts, while the input text was derived from a different utterance.

\textbf{Training.}
We trained {\pa} utilizing two A100 GPUs for 100 epochs with batch size of 6. We applied a learning rate of \( 1 \times 10^{-4} \) using the Adam optimizer with a 0.98 learning rate decay. The embedding size was set to 1024 for RVQ and 768 for the LM and SM. For all experiments, we used a random seed of 42 to ensure reproducibility. \newtext{For the overall generator loss, we select the weight coefficients proportionally as follows: 
\(\lambda_{L/LS} = X\), \(\lambda_f = 0.375X\), \(\lambda_t = 4.15X\), \(\lambda_w = 0.085X\), and set \(\lambda_g = 1\) and \(\lambda_{fm} = 1\).} \newtext{For DM-Codec-TTS, we trained the autoregressive (AR) and non-autoregressive (NAR) models separately, each for 4 epochs, while DM-Codec-TTS-small was trained with 100 epochs for the AR model and 196 epochs for the NAR model. The batch size was determined dynamically based on the maximum number of audio seconds, set to 280 for AR and 200 for NAR. We employed a base learning rate of 0.05 using the ScaledAdam optimizer with warmup steps of 200.}

\textbf{Baselines.}
\looseness=-1
We compared {\pa} with baselines: EnCodec (24 kHz) \citep{encodec}, SpeechTokenizer \citep{speechtokenizer}, FACodec (NaturalSpeech3) \citep{naturalspeech3}, \acltext{DAC (16 kHz) \citep{dac}, BigCodec \citep{bigcodec}, and Moshi \citep{moshi}}. We reproduced SpeechTokenizer using its official training code and \acltext{used official model checkpoints for the rest.}  Additionally, we compared DM-Codec-TTS with neural codec language models: USLM (from SpeechTokenizer \citep{speechtokenizer}) and VALL-E \citep{valle}. Since VALL-E and USLM's training codes and models are not open-source, we relied on reported results. We also used USLM (libri), an official model checkpoint trained on LibriTTS and shared on GitHub. All baseline model artifacts are open source and were used according to their licenses and intended research use.

\textbf{Evaluation Metrics.} To measure context preservation, we used Word Error Rate (WER) and Word Information Lost (WIL), with Whisper (whisper-medium) \citep{whisper} generating the transcriptions. \acltext{WER quantifies transcription errors, while WIL measures the loss of key information:  
\[
\text{WER} = \frac{S + D + I}{N}, \quad \text{WIL} = 1 - \frac{N - S - D}{N}
\]
where \( N \) is the total number of words in the reference, \( S \) is the number of correctly recognized words, \( D \) is the number of deletions, and \( I \) is the number of insertions.} Groundtruth WER and WIL scores were included to account for Whisper’s transcription errors.
We assess the acoustic and semantic preservation using ViSQOL (Virtual Speech Quality Objective Listener) \citep{visqol} measuring spectro-temporal similarity, Short-Time Objective Intelligibility (STOI) measuring short-time correlations, and \acltext{and Perceptual Evaluation of Speech Quality (PESQ)  measuring speech quality based on perceptual auditory models}. We measured the Mean Opinion Score (MOS) and Similarity Mean Opinion Score (SMOS) through human evaluations with 50 English-proficient participants. Evaluators rated anonymized samples on a 1-to-5 scale. MOS is used to evaluate the naturalness, intelligibility, and clarity of speech, while SMOS is used to measure the similarity to the prompt speaker's voice.\acltext{We calculated UTMOS \citep{utmos}, a neural network-based automatic MOS predictor to further complement subjective MOS.} Additionally, speaker similarity (Similarity) was quantified using cosine similarity between normalized speaker embeddings extracted with WavLM-TDNN \citep{Chen_2022}. Further details on human evaluations are in Appendix \ref{sec:human-eval}.

% %\vspace{-3pt}
\section{Experimental Results and Discussion}
% %\vspace{-5pt}
\label{sec:experiments}
% \looseness=-1

\looseness=-1
In this section, we present experimental results evaluating \textit{speech reconstruction} with {\pa} variants: {\pa} (LM) with LM-guided distillation, {\pa} (LM+SM) with combined LM and SM distillation, and {\pa} (CLS) with [CLS] token guided distillation against baseline speech tokenizers (\S\ref{subsec:speech-recons-eval}). We then assess zero-shot \textit{speech synthesis} with DM-Codec-TTS against neural codec language models (\S\ref{subsec:speech-syn-eval}). \acltext{Furthermore, we conducted rigorous \textit{significance analysis} (Appendix~\ref{ap-significance-analysis}) and comprehensive \textit{ablation studies} of DM-Codec design choices (Appendix~\ref{ap-ablation}).}

%%%%%%%%%%%%%%%%%%%%%%%%%%%%%%%%%%%%%%%%%%%%%%
%%%%%%%%%%%%%%%%%%%%%%%% START TABLE 2 %%%%%%%%%%%%%%%%%%%%%%%%%%%%%%%%%
\begin{table*}[hbt!]
\vspace{-3pt}
\caption{Evaluation of DM-Codec-TTS on LibriSpeech and VCTK datasets. Results show DM-Codec-TTS outperforms baselines in WER, WIL, Similarity, MOS, and SMOS. \textsuperscript{$\diamondsuit$} denotes results from official model checkpoints, \textsuperscript{$\dagger$} indicates results from the paper, and \textsuperscript{$\clubsuit$} indicates models trained with LibriTTS dataset. \textbf{Bold} highlights best result.}
\label{table:tts_eval}
\begin{center}
\resizebox{\textwidth}{!}{
\begin{tabular}{@{}l|ccccc|ccccc@{}}
\toprule
\multicolumn{1}{c|}{} & \multicolumn{5}{c|}{\textbf{LibriSpeech Benchmark Evaluation}} & \multicolumn{5}{c}{\textbf{VCTK Benchmark Evaluation}} \\ 
% \midrule 
\cmidrule(lr){2-6} \cmidrule(lr){7-11} 
\multicolumn{1}{c|}{\textbf{Model}} &
  \textbf{WER $\downarrow$} &
  \textbf{WIL $\downarrow$} &
  \textbf{Similarity $\uparrow$} &
  \textbf{MOS $\uparrow$} &
  \textbf{SMOS $\uparrow$} &  
  \textbf{WER $\downarrow$} &
  \textbf{WIL $\downarrow$} &
  \textbf{Similarity $\uparrow$} &
  \textbf{MOS $\uparrow$} &
  \textbf{SMOS $\uparrow$} \\ 
\midrule
DM-Codec-TTS & \textbf{5.08} & \textbf{7.32} & \textbf{0.82} & \textbf{3.70} & 3.89 & \textbf{3.58} & \textbf{5.65} & 0.82 & \textbf{3.78} & \textbf{3.85} \\
Vall-E \textsuperscript{$\dagger$} & 5.90 & - & 0.58 & - & \textbf{4.38} & - & - & 0.38 & - & 3.81 \\
USLM \textsuperscript{$\dagger$} &
  \multicolumn{1}{l}{} &
  - &
  - &
  - &
  - &
  6.50 &
  - &
  \textbf{0.84} &
  3.63 &
  3.45 \\
\midrule
DM-Codec-TTS\textsuperscript{$\clubsuit$} & 10.26 & 13.79 & 0.82 & 3.24 & 3.20 & 5.02 & 8.21 & 0.79 & 3.39 & 3.28 \\
\acltext{DM-Codec(SM)-TTS\textsuperscript{$\clubsuit$}} & \acltext{15.34} & \acltext{21.37} & \acltext{0.82} & \acltext{3.13} & \acltext{2.89} & \acltext{8.28} & \acltext{13.21} & \acltext{0.78} & \acltext{3.22} & \acltext{2.97} \\
USLM (libri) \textsuperscript{$\diamondsuit$} \textsuperscript{$\clubsuit$} & 16.72 & 25.65 & 0.80 & 3.11 & 2.83 & 14.79 & 23.24 & 0.78 & 2.94 & 2.63 \\
\bottomrule
\end{tabular}
}
\end{center}
\vspace{-10pt}
\end{table*}

%%%%%%%%%%%%%%%%%%%%%%%% START TABLE 2 %%%%%%%%%%%%%%%%%%%%%%%%%%%%%%%%%

%%%%%%%%%%%%%%%%%%%%%%%%%%%%%%%%%%%%%%%%%%%%%%

%\vspace{-3pt}
\subsection{Speech Reconstruction Evaluation}
%\vspace{-3pt}
\label{subsec:speech-recons-eval}

We compared the quality of {\pa}'s discrete speech representations by reconstructing speech from quantized features. \newtext{For this evaluation, we selected {\pa} variants with the first Residual Vector Quantizer layer (RVQ-1) for LM distillation and all RVQ layers (RVQ-1:8) for SM distillation.} \acltext{We analyze the results of DM-Codec in effectively incorporating semantic and contextual information.}

\textbf{Results:} 
Table~\ref{table:reconstruction1} shows that all {\pa} variants surpass or closely compete with the baselines. 
DM-Codec (LM+SM) outperforms all models in reducing transcription error (WER 4.05, WIL 6.61) and speech quality (ViSQOL 3.26, MOS 3.72, UTMOS 3.52). DM-Codec (LM) surpasses SpeechTokenizer, EnCodec, BigCodec, FACodec, and Mimi in reducing transcription error (WER 4.36, WIL 7.06) and speech quality (ViSQOL 3.18, PESQ 2.86), while achieving a higher UTMOS (3.69) than DAC, EnCodec, and Mimi. \acltext{Additionaly, DM-Codec (CLS) demonstrates commendable results, either outperforming or competing with the baselines, particularly outscoring SpeechTokenizer, EnCodec, BigCodec, FACodec, and Mimi in reducing transcription error (WER 4.47, WIL 7.08) and SpeechTokenizer, EnCodec, and Mimi in speech quality (ViSQOL 3.12, PESQ 2.65), while achieving a strong MOS of 3.65.}

% \looseness=1
\textbf{Discussion:} The superior performance of DM-Codec (LM+SM) is attributed to the novel Combined LM and SM-guided distillation. This dual representation, contextual knowledge from LM and semantic understanding from SM enables more coherent and natural speech, yielding superior results across most evaluation metrics. DM-Codec (LM) also performs well by incorporating contextual representations, improving speech quality and reducing transcription errors. \acltext{Additionally, [CLS] token-guided distillation provides a holistic view of the entire contextual input, enhancing contextual cue alignment and reducing transcription errors.} The impact of these distillation techniques is clear when compared to DM-Codec (Baseline), which lacks distillation and falls short in speech quality with higher transcription errors. Moreover, the importance of contextual information is reinforced by the comparison with DM-Codec (SM Baseline), which only includes semantic knowledge and underperforms with more transcription errors.

% %\vspace{-10pt}
\subsection{Speech Synthesis Evaluation}
\label{subsec:speech-syn-eval}
% %\vspace{-10pt}

\acltext{To demonstrate DM-Codec’s ability to capture semantic and contextual features and their impact on downstream task}, we compare the zero-shot TTS evaluation of DM-Codec-TTS with baseline USLM and VALL-E. We use DM-Codec (LM+SM) to quantize speech prompt features as input and decode quantized features predicted by DM-Codec-TTS. DM-Codec(SM)-TTS serves as a baseline.

\textbf{Results:} 
\newtext{The results in Table \ref{table:tts_eval} show that DM-Codec-TTS outperforms USLM and VALL-E baselines. In both benchmark evaluations, DM-Codec achieves the lowest WER (5.08, 3.58), WIL (7.32, 5.65), and highest MOS (3.70, 3.78), while achieving closely aligned Similarity to USLM in VCTK, and superior SMOS 3.85 compared to VALL-E and USLM in VCTK. DM-Codec-TTS trained with smaller LibriTTS dataset also significantly outperforms USLM (libri) in both benchmarks.} 

\textbf{Discussion:} \newtext{The improved performance indicates that hierarchical modeling in DM-Codec-TTS effectively utilizes the contextual and semantic knowledge distilled in DM-Codec. Unlike VALL-E, which relies on EnCodec for speech tokenization, DM-Codec-TTS leverages multimodal representation cues. This highlights the strength of contextual and semantic-aware hierarchical modeling in bridging linguistic content with acoustic fidelity, resulting in more natural and intelligible speech synthesis. \acltext{The improvement over the DM-Codec(SM)-TTS baseline also demonstrates the importance of contextual distillation from LM.}
}

\section{Conclusion}
% \vspace{-7pt}
We introduced {\pa}, a speech tokenizer with novel distillation methods that leverage multimodal (acoustic, semantic, and contextual) representations from language and speech self-supervised models. Experimental results demonstrate that distilling multimodal representations enables {\pa} to encode salient speech information in discrete tokens. This approach showcases the potential of multimodal representations to improve speech tokenization across domains, including multilingual and code-switched speech processing.

\section*{Limitations} In this work, we present the effectiveness of our proposed method, DM-Codec, based on the LibriSpeech dataset. Future research could investigate its performance across a variety of datasets and domains. Additionally, exploring the capabilities of DM-Codec in multilingual contexts would be valuable. Another limitation of our work is the absence of experiments with emerging LLMs. Currently, we focus solely on masked language models to derive representations. Further investigation into decoder-based LLMs' impact on DM-Codec can be studied and addressed.

\section*{Acknowledgments}
This project was partially supported by a grant from Independent University, Bangladesh (IUB).

% Bibliography entries for the entire Anthology, followed by custom entries
%\bibliography{anthology,custom}
% Custom bibliography entries only
\bibliography{main}

\appendix
\clearpage
\begin{center}
{\Large \textbf{Technical Appendix}}
% \\DM-Codec: \textbf{\underline{D}}istilling \textbf{\underline{M}}ultimodal\\ Representations for Speech Tokenization}    
\end{center}

%%%%%%%%%%%%%%%%%%%%%%%%%%%%%%%%%%%%%%%
\section{Resources}
\label{sec:resource}
%%%%%%%%%%%%%%%%%%%%%%%%%%%%%%%%%%%%%%%
We release all resources to facilitate reproducibility and future research. We provide the code for training DM-Codec, trained model checkpoints for inference, and a Dockerfile for a reproducible environment. All resources are publicly available at \href{https://github.com/mubtasimahasan/DM-Codec}{https://github.com/mubtasimahasan/DM-Codec}.

% %%%%%%%%%%%%%%%%%%%%%%%%%%%%%%%%%%%%%%%
% \input{sections/related_work}
% %%%%%%%%%%%%%%%%%%%%%%%%%%%%%%%%%%%%%%%

%%%%%%%%%%???????????????%%%%%%%%%%%%%%
%%%%%%% To make it appear earlier
%%%%%%%%%%%%%%%%%%%%%%% START TABLE 2 %%%%%%%%%%%%%%%%%%%%%%%%%%%%
% \vspace{-3pt}
\begin{table*}[t]
\centering
\caption{\newtext{Significance Analysis on LibriSpeech Test Set.} Significance analysis is conducted at $\alpha = 0.05$ between LM and SM-guided \textbf{DM-Codec (D), EnCodec (E), SpeechTokenizer (S), and FACodec (F)}. Comparisons are performed row vs. column (e.g., D vs. E, E vs. S). Results reveal DM-Codec consistently achieves significantly better scores in key metrics across all individual samples. A \cmark\ indicates significance, a \bmark\ denotes dominance, and a \xmark\ means no significance. Avg and Std mean the average and standard deviation of each score.}
\label{table:significance}
\resizebox{\textwidth}{!}{ 
\begin{tabular}{@{}l|cccccc|cccccc|cccccc|cccccc@{}}
\toprule
                & \multicolumn{6}{c|}{\textbf{WER $\downarrow$}}       & \multicolumn{6}{c|}{\textbf{WIL $\downarrow$}}       & \multicolumn{6}{c|}{\textbf{ViSQOL $\uparrow$}}    & \multicolumn{6}{c}{\textbf{STOI $\uparrow$}}      \\ \midrule
 & \textbf{Avg} & \textbf{Std} & \textbf{D} & \textbf{E} & \textbf{S} & \textbf{F} & \textbf{Avg} & \textbf{Std} & \textbf{D} & \textbf{E} & \textbf{S} & \textbf{F} & \textbf{Avg} & \textbf{Std} & \textbf{D} & \textbf{E} & \textbf{S} & \textbf{F} & \textbf{Avg} & \textbf{Std} & \textbf{D} & \textbf{E} & \textbf{S} & \textbf{F} \\ \midrule
\textbf{D}     & 4.774        & 0.100        & -& \cmark       & \cmark       & \cmark       & 7.510        & 0.139        & -& \cmark       & \cmark       & \cmark       & 3.197        & 0.184        & -& \bmark       & \cmark       & \cmark       & 0.937        & 0.021        & -& \cmark       & \cmark       & \xmark       \\ \midrule
\textbf{E}        & 4.828        & 0.100        & \xmark       & -& \cmark       & \cmark       & 7.593        & 0.137        & \xmark       & -& \cmark       & \cmark       & 3.064        & 0.201        & \xmark       & -& \xmark       & \xmark       & 0.917        & 0.021        & \xmark       & -& \xmark       & \xmark       \\ \midrule
\textbf{S}  & 4.942        & 0.101        & \xmark       & \xmark       & -& \xmark       & 7.725        & 0.138        & \xmark       & \xmark       & -& \xmark       & 3.080        & 0.190        & \xmark       & \cmark       & -& \xmark       & 0.920        & 0.025        & \xmark       & \cmark       & -& \xmark       \\ \midrule
\textbf{F}       & 4.914        & 0.103        & \xmark       & \xmark       & \cmark       & -& 7.643        & 0.141        & \xmark       & \xmark       & \cmark       & -& 3.113        & 0.250        & \xmark       & \cmark       & \cmark       & -& 0.946        & 0.023        & \cmark       & \cmark       & \cmark       & -\\ 

\bottomrule
\end{tabular}}
% \vspace{-.5cm}
\end{table*}
%%%%%%%%%%%%%%%%%%%%%%% End TABLE 2 %%%%%%%%%%%%%%%%%%%%%%%%%%%%%%
%%%%%%%%%%???????????????%%%%%%%%%%%%%%

\section{\newtext{Representation Alignment in LM Distillation}}

\newtext{A core principle of our approach is that strict temporal alignment between text and acoustic representations is not necessary for effective contextual knowledge distillation. The discrete representations produced by the Residual Vector Quantizer (RVQ) inherently encode holistic information about the speech segment rather than temporally localized features. This characteristic enables the distillation of contextual representations into discrete tokens without reliance on strict temporal correspondence.}

\newtext{As illustrated in Figure \ref{fig:tokenizer}, the RVQ outputs \(\{Q_1, \dots, Q_{T^{\prime}}\}\) capture comprehensive speech information across the entire utterance. Temporal alignment between these discrete representations and contextual representations (e.g., from BERT) or semantic representations (e.g., from HuBERT) is therefore not essential. The effectiveness of dimension-level semantic representation distillation without temporal alignment has been previously demonstrated in SpeechTokenizer \citep{speechtokenizer}, providing a robust theoretical foundation for this approach.}

\newtext{Moreover, vector quantizer output representations are not inherently aligned with time steps, but instead encode holistic speech information. This capability has been established in prior work \citep{speechtokenizer, priorwork1, priorwork2, priorwork3}, which demonstrates the potential of vector quantization to discretize input data into intermediate representations that capture essential features across the feature dimension. Consequently, imposing a temporal alignment between vector quantizer outputs and hidden layer representations from language or semantic models would neither align with the methodological objectives nor enhance the efficacy of the proposed distillation approach.}

\newtext{To achieve effective contextual and semantic knowledge transfer, we employ a continuous distillation loss that \textbf{maximizes cosine similarity at the feature dimension level} between the selected RVQ layer outputs and the teacher representations across all time steps. Unlike conventional methods that rely on time-step-wise loss calculations \citep{speechtokenizer, timesteploss}, this dimension-level cosine similarity loss ensures that DM-Codec captures contextual and semantic knowledge through LM-Guided Distillation and Combined LM and SM-Guided Distillation mechanisms, without requiring strict temporal alignment.}

\newtext{In addition, we propose a \textbf{[CLS]-token-based distillation strategy} to address alignment. The [CLS] token encodes sequence-level holistic representations, capturing global contextual information from language models. Using these token representations, our method eliminates the need for fine-grained temporal alignment while preserving essential linguistic features. This complements the dimension-level distillation strategy by focusing on global sequence features, enabling the adaptability of our approach to scenarios where fine-grained alignment is infeasible or unnecessary.}

\newtext{As shown in Table \ref{table:reconstruction1}, all DM-Codec variants, including DM-Codec (CLS), DM-Codec (LM), and DM-Codec (LM+SM) consistently outperform baseline models (EnCodec, SpeechTokenizer, and FACodec) in content preservation metrics (WER and WIL) while maintaining competitive performance in speech quality metrics (ViSQOL and STOI). These results corroborate the robustness of the proposed approach.} \newtext{By prioritizing the holistic integration of multimodal knowledge into discrete speech representations, DM-Codec achieves significant advancements in both content preservation and speech quality. }

%%%%%%%%%%%%%%%%%%%%%%%%%%%%%%%%%%%%%%%
\section{Significance Analysis of Speech Tokenizer performance}
\label{ap-significance-analysis}

% \vspace{-5pt}
We conducted a significance analysis at \( \alpha = 0.05 \) on the LibriSpeech test-clean subset containing 2,620 samples. We follow the approach of \citet{Dior}, to measure the stochastic dominance of {\pa} over the baselines: EnCodec, SpeechTokenizer, and FACodec. Specifically, we computed inverse cumulative distribution functions (CDFs) for all reconstructed speech samples' individual WER, WIL, ViSQOL, and STOI scores. Significance was evaluated using the \( \epsilon \) value and categorized as: significantly better when \( 0.0 < \epsilon \leq 0.5 \), significantly dominant when \( \epsilon = 0.0 \), and not significantly better when \( \epsilon > 0.5 \). For this analysis, we selected {\pa} (LM+SM), trained with combined LM and SM-guided distillation. \textbf{To the best of our knowledge, we are the first to conduct significance analysis to measure the effectiveness of different speech tokenizers.} 

\textbf{Results and Discussion:} The results in Table \ref{table:significance} show that {\pa} significantly outperforms the baselines in WER, WIL, ViSQOL, and STOI scores. The improved average values (4.774 WER, 7.510 WIL, 3.197 ViSQOL, 0.937 STOI) and consistent standard deviations (0.100 WER, 0.139 WIL, 0.184 ViSQOL, 0.021 STOI) further demonstrate the statistical significance. Notably, {\pa}'s performance in WER and WIL underscores the importance of contextual representation distillation for enhanced speech reconstruction. Additionally, its strong performance in ViSQOL and STOI, especially over EnCodec, highlights the benefits of combining LM and SM distillation for retaining semantic-acoustic fidelity. While {\pa} does not achieve significance over FACodec in terms of STOI, it significantly outperforms the baselines across all other metrics. Among the baseline, FACodec achieves significance over SpeechTokenizer, whereas EnCodec outperforms SpeechTokenizer in WER and WIL, SpeechTokenizer excels in ViSQOL and STOI over EnCodec.

%%%%%%%%%%%%%%%%%%%%%%%%%%%%%%%%%%%%%%%

%%%%%%%%%%???????????????%%%%%%%%%%%%%%
%%%%%%% To make it appear earlier
%%%%%%%%%%%%%%%%%%%%%%%% START Figure: Distil %%%%%%%%%%%%%%%%%%
\begin{figure*}[t]
\centering
\begin{minipage}{0.72\textwidth}
    \includegraphics[width=\textwidth]{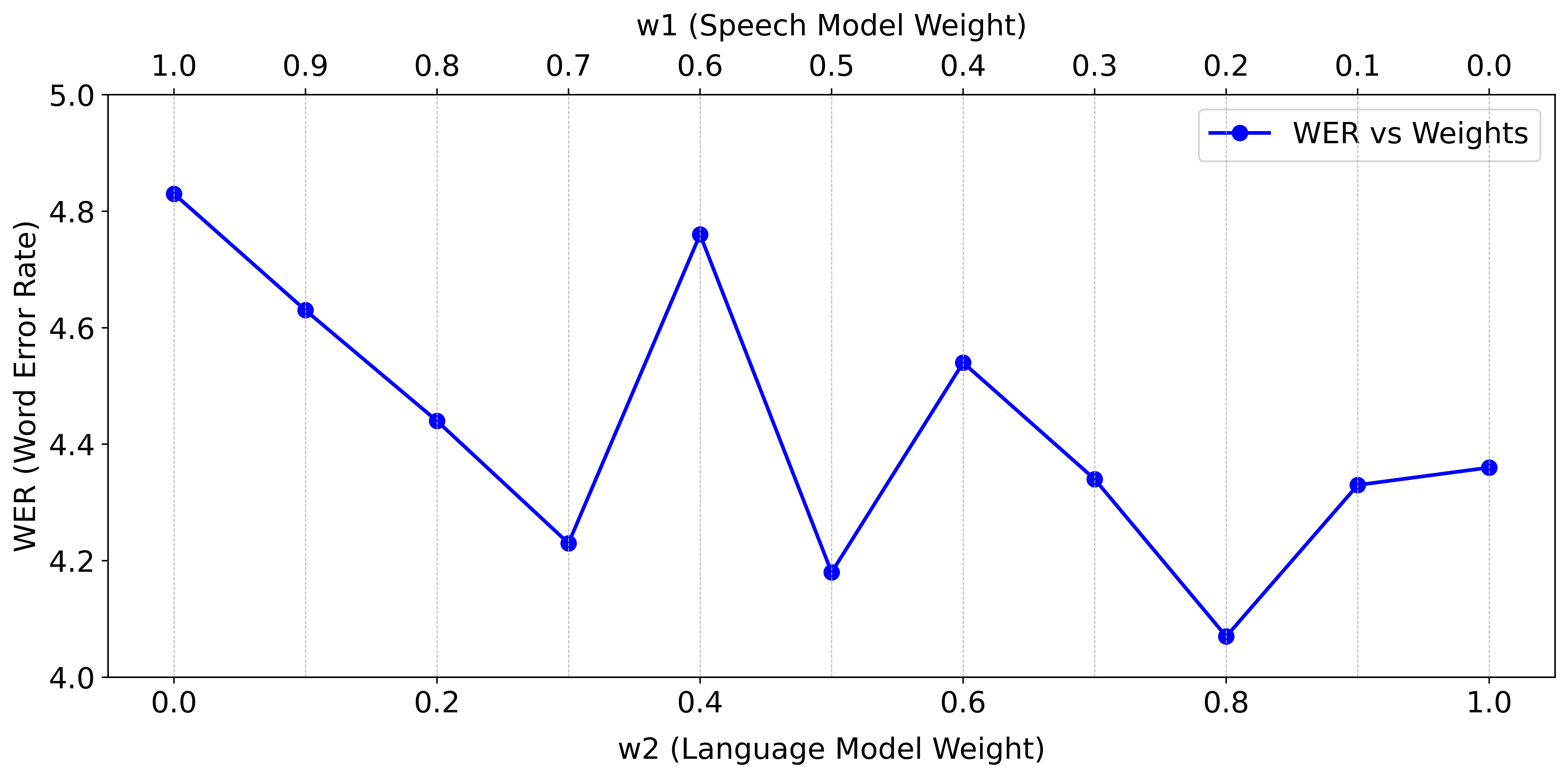}
\end{minipage}
\hfill 
\begin{minipage}{0.22\textwidth} 
    \begin{center}
\resizebox{0.9\textwidth}{!}{ 
    \begin{tabular}{ccc}
    \toprule
    % \hline
    \multicolumn{1}{c}{\(\lambda_{SM}\)} & \multicolumn{1}{c}{\(\lambda_{LM}\)} & \multicolumn{1}{c}{\bf WER $\downarrow$} \\
    % \hline
    \midrule
    1.0 & 0.0 & 4.83 \\
    0.9 & 0.1 & 4.63 \\
    0.8 & 0.2 & 4.44 \\
    0.7 & 0.3 & 4.23 \\
    0.6 & 0.4 & 4.76 \\
    0.5 & 0.5 & 4.18 \\
    0.4 & 0.6 & 4.54 \\
    0.3 & 0.7 & 4.34 \\
    0.2 & 0.8 & 4.07 \\
    0.1 & 0.9 & 4.33 \\
    0.0 & 1.0 & 4.36 \\
    % \hline
    \bottomrule
    \end{tabular}
}    
    \end{center}
\end{minipage}
\caption{Effects of weights on combined distillation from Speech Model (SM) and Language Model (LM). Higher LM weight generally results in improved WER, suggesting its stronger contribution to content preservation. Here, \(\lambda_{SM}\) is the weight added to SM and \(\lambda_{LM}\) is the weight added to LM, where (\(\lambda_{SM}\) + \(\lambda_{LM}\) = 1).}
\label{fig:weights}
\end{figure*}
%%%%%%%%%%%%%%%%%%%%%% START Table: Layers %%%%%%%%%%%%%%%%%%%%

%%%%%%%%%%???????????????%%%%%%%%%%%%%%

%%%%%%%%%%%%%%%%%%%%%%%%%%%%%%%%%%%%%%%
% \input{tables/ab_weights} %%%%%%%%%%%%%%%%%% Its in previous section

\section{Ablation Studies}
\label{ap-ablation}
\acltext{We conducted a comprehensive analysis of {\pa}'s performance and the impact of each methodological choice in retaining contextual and semantic information through distillation.} Unless otherwise stated, we use distillation for both LM and SM to the first Residual Vector Quantizer layer (RVQ-1) for comparison consistency and simplicity. \newtext{The following ablation studies were conducted concurrently with a similar configuration and model design except for the explicitly noted changes.}

\subsection{Ablation Study: Impact of Combined Semantic Distillation}
\label{sec:weights}

% \looseness=-1
We conducted experiments with different weighted combinations of LM and SM distillation loss to evaluate their impact on reducing WER. The combined distillation loss from Equation \ref{eq:distil} was updated using SM and LM weights ($\lambda_{SM}$ and $\lambda_{LM}$), ranging from $0.0$ to $1.0$, with the constraint $\lambda_{SM} + \lambda_{LM} = 1$.

% % \vspace{-.2cm}
\begin{equation}
\mathcal{L}_{LS} = \frac{1}{2} \left( \lambda_{LM} \cdot \mathcal{L}_{L}+\lambda_{SM} \cdot \mathcal{L}_{S}  \right)
\end{equation}
% % \vspace{-.2cm}

\looseness=-1
\textbf{Results and Discussion:} The experimental results are presented in Figure~\ref{fig:weights}, showing the speech reconstruction results with WER scores for different weighted combinations. From the values, we notice a trend showing that incorporating LM representations generally improves WER, especially when LM distillation is dominant. The lowest WER score of 4.07 occurs with a weight of \(\lambda_{LM} = 0.8\) for LM, while \(\lambda_{SM} = 0.2\) for SM, highlighting the strong influence of LM distillation on capturing contextual information. A balanced weighting of \(\lambda_{SM} = 0.5\) and \(\lambda_{LM} = 0.5\) produces a WER of 4.18, confirming that distillation from both LM and SM is beneficial. However, as the weighting shifts more in favor of SM (\(\lambda_{SM} > 0.7\)), WER deteriorates, reaching 4.83 when relying entirely on SM. This underscores that over-reliance on SM distillation compromises contextual accuracy in favor of raw speech features. \newtext{Notably, the interaction between LM and SM weights plays a crucial role, as the combined distillation influences the overall WER beyond individual distillation contributions. For instance, the higher WER observed at \(\lambda_{LM} = 0.9\) compared to \(\lambda_{LM} = 0.3\) or \(\lambda_{LM} = 0.5\) highlights the importance of tuning both weights synergistically, rather than favoring one in isolation.}

%%%%%%%%%%%%%%%%%%%%%%%%%%%%%%%%
%%%%%%%%%%%%%%%%%%%%%%%% START Table: Layer %%%%%%%%%%%%%%%%%%%%%%%%%%%%
\begin{table*}[t]
\caption{Analysis of different RVQ layers effect on speech reconstruction. LM-guided distillation on RVQ-1 layer ensures greater content preservation, while SM-guided distillation on RVQ-1:8 layer is more effective at preserving semantic representation. LM-layer and SM-layer indicate the RVQ layer used for respective distillation. 
(LM) indicates LM-guided Distillation. (LM+SM) indicates combined LM and SM-guided Distillation. \textbf{Bold} highlights the best result and \underline{underline} the second-best result.}
\label{table:layers}
\begin{center}
\small
\begin{tabular}{l cccccc}

\toprule
\multicolumn{1}{c}{\bf Tokenizer} & \multicolumn{1}{c}{\bf LM-Layer} & \multicolumn{1}{c}{\bf SM-Layer} & \multicolumn{1}{c}{\bf WER $\downarrow$} & \multicolumn{1}{c}{\bf WIL $\downarrow$} & \multicolumn{1}{c}{\bf ViSQOL $\uparrow$} & \multicolumn{1}{c}{\bf STOI $\uparrow$} \\

\midrule
{\pa} (LM) & RVQ-1 & - & 4.36 & 7.06 & 3.18 & 0.935 \\ 
{\pa} (LM) & RVQ-1:8 & - & 4.23 & 6.94 & 3.12 & 0.929 \\ 
{\pa} (LM) & RVQ-8 & - & 4.44 & 7.22 & 3.28 & 0.935 \\ 

\midrule
{\pa} (LM+SM) & RVQ-1 & RVQ-1 & \underline{4.18} & \underline{6.84} & 3.13 & 0.933 \\ 
{\pa} (LM+SM) & RVQ-1:8 & RVQ-1 & 4.59 & 7.34 & 3.21 & 0.937 \\ 
{\pa} (LM+SM) & RVQ-8 & RVQ-1 & 4.49 & 7.24 & \underline{3.30} & \underline{0.938} \\ 
{\pa} (LM+SM) & RVQ-1 & RVQ-1:8 & \textbf{4.05} & \textbf{6.61} & 3.26 & 0.937 \\ 
{\pa} (LM+SM) & RVQ-1 & RVQ-8 & 4.39 & 7.08 & \textbf{3.33} & \textbf{0.939} \\ 

\bottomrule
\end{tabular}
% }
\end{center}
\end{table*}
% % \vspace{-1cm}
%%%%%%%%%%%%%%%%%%%%%%%% START Table: Layer %%%%%%%%%%%%%%%%%%%%%%%%%%%%
%%%%%%%%%%%%%%%%%%%%%%%%%%%%%%%%

\subsection{Ablation Study: Impact of Distillation on Different RVQ Layers}
\label{sec:layers}

We evaluated the effect of applying distillation at various Residual Vector Quantizer (RVQ) layers, including the first layer (RVQ-1), the average of eight layers (RVQ-1:8), and the last layer (RVQ-8). Table \ref{table:layers} shows the full results.

\textbf{Results and Discussion:} In LM-guided distillation, RVQ-1:8 achieves the best WER and WIL scores (4.23 and 6.94), though with lower ViSQOL and STOI scores (3.12 and 0.929) compared to RVQ-8 (3.28 and 0.935). The RVQ-1 layer provides the best overall balance between content preservation and perceptual quality, with WER, WIL, ViSQOL, and STOI scores of 4.36, 7.06, 3.18, and 0.935. This demonstrates RVQ-1:8 prioritizes contextual integrity, while RVQ-8 favors perceptual quality. Thus, we select RVQ-1 for LM-guided distillation due to its balanced performance.

For LM and SM-based distillation, the RVQ-1 and RVQ-1:8 combination achieves the best WER and WIL scores (4.05 and 6.61), with RVQ-1 and RVQ-1 as the second-best (4.18 and 6.84). In contrast, the RVQ-1 and RVQ-8 combination yields the highest ViSQOL and STOI scores (3.33 and 0.939), followed by RVQ-8 and RVQ-1 (3.30 and 0.938). RVQ-1 captures contextual representation more effectively due to its simpler quantized vector, while RVQ-1:8 incorporates more nuanced semantic and acoustic aspects. Overall, this ablation shows that selecting RVQ layers for LM and SM-based distillation greatly affects the balance between contextual accuracy and semantic-acoustic fidelity, allowing layer combinations to be tailored to task requirements.

%%%%%%%%%%%%%%%%%%%%%%%%%%%%%%%%
%%%%%%%%%%%%%%%%%%%%%%%% START Table: Models %%%%%%%%%%%%%%%%%%%%%%%%%%%%
\begin{table*}[t]
\caption{Analysis of representation distillation from different models. BERT can be effectively combined with HuBERT or wav2vec 2.0, however, ELECTRA in LM-guided distillation outperforms BERT. (LM) indicates LM-guided Distillation. (LM+SM) indicates combined LM and SM-guided Distillation. \textbf{Bold} highlights the best result and \underline{underline} the second-best result.}

\label{table:models}
\begin{center}
\small
\begin{tabular}{l cccccc}
% \hline
\toprule
\multicolumn{1}{c}{\bf Tokenizer} & \multicolumn{1}{c}{\bf LM} & \multicolumn{1}{c}{\bf SM} & \multicolumn{1}{c}{\bf WER $\downarrow$} & \multicolumn{1}{c}{\bf WIL $\downarrow$} & \multicolumn{1}{c}{\bf ViSQOL $\uparrow$} & \multicolumn{1}{c}{\bf STOI $\uparrow$} \\

% \hline 
\midrule
{\pa} (LM) & BERT & - & 4.36 & 7.06 & \underline{3.18} & 0.935 \\ 
{\pa} (LM) & ELECTRA & - & \textbf{4.12} & \textbf{6.63} & 3.10 & \underline{0.936} \\ 
% \hline
\midrule
{\pa} (LM+SM) & BERT & HuBERT & 4.18 & 6.84 & 3.13 & 0.933 \\
{\pa} (LM+SM) & BERT & wav2vec 2.0 & \underline{4.13} & \underline{6.77} & \textbf{3.15} & \textbf{0.942} \\ 
{\pa} (LM+SM) & ELECTRA & wav2vec 2.0 & 4.70 & 7.51 & 3.14 & 0.933 \\ 
{\pa} (LM+SM) & ELECTRA & HuBERT & 4.67 & 7.58 & 2.94 & 0.932 \\
% \hline
\bottomrule
\end{tabular}
% }
\end{center}
\end{table*}
%%%%%%%%%%%%%%%%%%%%%%%% END Table: Models %%%%%%%%%%%%%%%%%%%%%%%%%%%%
%%%%%%%%%%%%%%%%%%%%%%%%%%%%%%%%

\subsection{Ablation Study: Impact of Different Models on Distillation}

We experimented with different LM and SM distillations to analyze performance variations based on different model selections. In addition to our selected BERT \citep{bert} and HuBERT \citep{hubert}, we experiment with ELECTRA (electra-base-discriminator) \citep{electra} as the LM and wav2vec 2.0 (wav2vec2-base-960h) \citep{wav2vec20} as the SM. Table \ref{table:models} shows the full results.

\textbf{Results and Discussion:} In LM-guided distillation, the ELECTRA model significantly enhances performance, achieving WER and WIL scores of 4.12 and 6.63, respectively, compared to BERT's scores of 4.36 and 7.06. This indicates the architecture of ELECTRA's effectiveness for the proposed LM-guided distillation, demonstrating its superior contextual representation. These results are consistent with ELECTRA's better performance in general natural language processing tasks. However, we select BERT for its simplicity and established performance. 

In LM and SM-guided distillation, the combination of BERT and wav2vec 2.0 achieves the highest overall performance, with scores of WER 4.13, WIL 6.77, ViSQOL 3.15, and STOI 0.942. However, the combination of BERT and HuBERT closely follows with second-best scores of WER 4.18, WIL 6.84, and ViSQOL 0.933. These findings demonstrate that different speech models can be effectively integrated with the BERT model.

%%%%%%%%%%%%%%%%%%%%%%%%%%%%%%%%
%%%%%%%%%%%%%%%%%%%%%%%% START Table: Semantic %%%%%%%%%%%%%%%%%%%%%%%%%%%
\begin{table*}[t]
\caption{Analysis of different distillation layers representation on speech reconstruction. Average layer provides more comprehensive representations. (LM) indicates LM-guided Distillation. (LM+SM) indicates combined LM and SM-guided Distillation. \textbf{Bold} highlights the best result and \underline{underline} the second-best result.}

\label{table:semantic}
\begin{center}
\small 
\begin{tabular}{l cccccc}
% \hline
\toprule
\multicolumn{1}{c}{\bf Tokenizer} & \multicolumn{1}{c}{\bf Distillation Layer(s)} & \multicolumn{1}{c}{\bf WER $\downarrow$} & \multicolumn{1}{c}{\bf WIL $\downarrow$} & \multicolumn{1}{c}{\bf ViSQOL $\uparrow$} & \multicolumn{1}{c}{\bf STOI $\uparrow$} \\

% \hline 
\midrule
{\pa} (LM) & Average & \underline{4.36} & \underline{7.06} & \textbf{3.18} & \textbf{0.935} \\ 
{\pa} (LM) & Last & 4.62 & 7.56 & 2.95 & 0.926 \\ 
{\pa} (LM) & $9^{\text{th}}$ & 4.75 & 7.80 & 2.88 & 0.925 \\ 
% \hline 
\midrule
{\pa} (LM+SM) & Average & \textbf{4.18} & \textbf{6.84} & \underline{3.13} & \underline{0.933} \\
{\pa} (LM+SM) & Last & 4.68 & 7.55 & 3.03 & 0.933 \\
{\pa} (LM+SM) & $9^{\text{th}}$ & 4.52 & 7.43 & 3.00
& 0.933 \\
% \hline
\bottomrule
\end{tabular}
% } % End of resizebox
\end{center}
\end{table*}
%%%%%%%%%%%%%%%%%%%%%%%% END Table: Semantic %%%%%%%%%%%%%%%%%%%%%%%%%%%
%%%%%%%%%%%%%%%%%%%%%%%%%%%%%%%%

%%%%%%%%%%%%%%%%%%%%%%%%%%%%%%%%
%%%%%%%%%%%%%%%%%%%%%%%% START TABLE D %%%%%%%%%%%%%%%%%%%%%%%%%%%%%%%%%
\begin{table*}[t]
\caption{\newtext{Analysis of different bit rates for speech reconstruction. {\pa} (LM+SM) showcases its robustness at reduced bitrates, outperforming baselines at 3 kbps and maintaining competitive content preservation scores (WER, WIL) and superior speech quality (ViSQOL, STOI) at 1.5 kbps. \newtext{$f_s$ represents the audio sample rate, and $f_r$ the codec frame rate.} \textsuperscript{$\heartsuit$} means the results were reproduced using the official training code. \textsuperscript{$\diamondsuit$} means the results were obtained using official model checkpoints. (LM) indicates LM-guided Distillation method. (LM+SM) indicates combined LM and SM-guided Distillation method.}}
\label{table:low_bitrate}
\begin{center}
\small
\begin{tabular}{l cccc ccc}

\toprule
\multicolumn{1}{c}{\bf Model} & \multicolumn{1}{c}{$f_s$} & \multicolumn{1}{c}{$f_r$} & \multicolumn{1}{c}{\bf Bitrate} & \multicolumn{1}{c}{\bf WER $\downarrow$} & \multicolumn{1}{c}{\bf WIL $\downarrow$} & \multicolumn{1}{c}{\bf ViSQOL $\uparrow$} & \multicolumn{1}{c}{\bf STOI $\uparrow$} \\

\midrule
{\pa} (LM+SM) & 16 kHz & 50 Hz & 3 kbps & \textbf{4.29} & \textbf{7.04} & \textbf{3.070} & \textbf{0.928} \\ 
{\pa} (LM) & 16 kHz & 50 Hz & 3 kbps & 4.38 & 7.09 & 3.042 & 0.924 \\ 
SpeechTokenizer\textsuperscript{$\heartsuit$} & 16 kHz & 50 Hz & 3 kbps & 4.70 & 7.43 & 2.905 & 0.911 \\ 
EnCodec\textsuperscript{$\diamondsuit$} & 24 kHz & 50 Hz & 3 kbps & 4.80 & 7.80 & 2.550 & 0.872 \\ 

\midrule
{\pa} (LM) & 16 kHz & 50 Hz & 1.5 kbps & 6.14 & 10.13 & 2.644 & 0.880 \\ 
{\pa} (LM+SM) & 16 kHz & 50 Hz & 1.5 kbps & 6.19 & 10.16 & \textbf{2.662} & \textbf{0.894} \\ 
SpeechTokenizer\textsuperscript{$\heartsuit$} & 16 kHz & 50 Hz & 1.5 kbps & \textbf{5.61} & \textbf{9.02} & 2.500 & 0.846 \\ 
EnCodec\textsuperscript{$\diamondsuit$} & 24 kHz & 50 Hz & 1.5 kbps & 10.53 & 16.63 & 2.443 & 0.809 \\ 

\midrule
\acltext{DM-Codec (LM)} & \acltext{16 kHz} & \acltext{50 Hz} & \acltext{0.75 kbps} & \acltext{7.96} & \acltext{12.65} & \acltext{2.18} & \acltext{0.583} \\ 
\acltext{DM-Codec (LM+SM)} & \acltext{16 kHz} & \acltext{50 Hz} & \acltext{0.75 kbps} & \acltext{7.81} & \acltext{12.42} & \acltext{\textbf{2.20}} & \acltext{\textbf{0.675}} \\ 
\acltext{SpeechTokenizer\textsuperscript{$\heartsuit$}} & \acltext{16 kHz} & \acltext{50 Hz} & \acltext{0.75 kbps} & \acltext{\textbf{7.08}} & \acltext{\textbf{11.67}} & \acltext{2.07} & \acltext{0.556} \\ 

\bottomrule
\end{tabular}
% } % End of resizebox
\end{center}
% \vspace{-.5cm}
\end{table*}
%%%%%%%%%%%%%%%%%%%%%%%% End TABLE D %%%%%%%%%%%%%%%%%%%%%%%%%%%%%%%%%
%%%%%%%%%%%%%%%%%%%%%%%%%%%%%%%%

\subsection{Ablation Study: Impact of Different Distillation Layer(s)}

We evaluated speech reconstruction using different distillation layers of the LM and SM, examining which combination of layers yields the most relevant representations of semantic and contextual information. For this ablation, we considered the average of all layer representations, the $9^{\text{th}}$ layer representations, and the last layer representations. Table \ref{table:semantic} shows the full results.

\textbf{Results and Discussion:}  In LM-guided distillation, the use of the average layer achieves superior overall performance, with a WER of 4.36, WIL of 7.06, ViSQOL of 3.18, and STOI of 0.935, compared to the variants utilizing the last and 9th layers. Similarly, in LM and SM-guided distillation, the average layer yields superior results compared to the last and $9^{\text{th}}$ layer variants.

The results indicate that averaging all layers leads to more comprehensive representations of semantic or contextual information. In the case of LM, the averaging process provides greater contextual representation and synergizes syntactic information from earlier layers and abstract word relations from higher layers. In combined LM and SM-guided distillation, averaging all SM layers provides a more nuanced understanding of the earlier layer's phonetic information and the higher layers' richer semantic information. Conversely, relying solely on the last layer or the $9^{\text{th}}$ layer fails to capture the overall context and semantic information, yielding less relevant representation distillation.

\subsection{\newtext{Ablation Study: Impact of Low Bit Rate}}

\newtext{We evaluated zero-shot speech reconstruction at reduced bitrates of 3kbps, 1.5kbps, and 0.75kbps. DM-Codec and the baseline SpeechTokenizer were trained on 16kHz sample rates, whereas the EnCodec baseline was trained on 24kHz sample rates. To achieve lower bitrates in DM-Codec and SpeechTokenizer, we limited RVQ levels: the first 6 RVQ layers for 3kbps, 3 RVQ layers for 1.5kbps, and 1 RVQ layer for 0.75kbps. For EnCodec, we kept the first 4 VQ layers for 3kbps and 2 VQ layers for 1.5kbps, and it does not support a 0.75kbps bitrate. In this ablation, {\pa} variants used the first RVQ layer (RVQ-1) for LM distillation, while all RVQ layers (RVQ-1:8) were used for SM distillation. Table \ref{table:low_bitrate} presents the full results.}

\newtext{\textbf{Results and Discussion:} At a 3kbps bitrate, LM-guided DM-Codec (LM) maintains its performance and consistency, surpassing the baseline with scores of 4.38 WER, 7.09 WIL, 3.042 ViSQOL, and 0.924 STOI. Combined LM and SM-guided DM-Codec (LM+SM) further improves these scores to 4.29 WER, 7.04 WIL, 3.070 ViSQOL, and 0.928 STOI, outperforming all baselines.}

\acltext{At 1.5kbps and 0.75kbps, DM-Codec (LM) and DM-Codec (LM+SM) achieve scores comparable to SpeechTokenizer in WER and WIL but maintain superior speaker quality, achieving the best ViSQOL and STOI scores of 2.662 and 0.894 at 1.5kbps, and 2.20 and 0.675 at 0.75kbps, respectively.}
\acltext{We hypothesize that the slight performance degradation in WER and WIL at lower bitrates is due to the loss of contextual representation, as the reduced bandwidth limits the model's ability to fully utilize nuanced contextual details incorporated into DM-Codec through distillation.}

%%%%%%%%%%%%%%%%%%%%%%%%%%%%%%%%
%%%%%%%%%%%%%%%%%%%%%%%% START TABLE E %%%%%%%%%%%%%%%%%%%%%%%%%%%%%%%%%
\begin{table*}[t]
\caption{\acltext{Analysis of different distillation axes for speech reconstruction. Distillation at the feature dimension is more robust and enhances reconstruction. D-axis indicates the proposed distillation along the feature dimension. T-axis indicates distillation along the time axis. (LM) indicates LM-guided Distillation. (LM+SM) indicates combined LM and SM-guided Distillation. \textbf{Bold} highlights the best result and \underline{underline} the second-best result.}}
\label{table:axis}
\begin{center}
\small
\begin{tabular}{l cccccc}
\toprule
\multicolumn{1}{c}{\bf \acltext{Model}} & \multicolumn{1}{c}{\bf \acltext{Axis}} & \multicolumn{1}{c}{\bf \acltext{WER $\downarrow$}} & \multicolumn{1}{c}{\bf \acltext{WIL $\downarrow$}} & \multicolumn{1}{c}{\bf \acltext{ViSQOL $\uparrow$}} & \multicolumn{1}{c}{\bf \acltext{STOI $\uparrow$}} \\
\midrule
\acltext{DM-Codec (LM)} & \acltext{D-axis} & \acltext{\underline{4.36}} & \acltext{\underline{7.06}} & \acltext{\textbf{3.18}} & \acltext{\textbf{0.935}} \\ 
\acltext{DM-Codec (LM)} & \acltext{T-axis} & \acltext{4.76} & \acltext{7.82} & \acltext{2.92} & \acltext{0.925} \\ 
\midrule
\acltext{DM-Codec (LM+SM)} & \acltext{D-axis} & \acltext{\textbf{4.05}} & \acltext{\textbf{6.61}} & \acltext{\underline{3.26}} & \acltext{\underline{0.937}} \\
\acltext{DM-Codec (LM+SM)} & \acltext{T-axis} & \acltext{4.76} & \acltext{7.76} & \acltext{2.86} & \acltext{0.926} \\
\bottomrule
\end{tabular}
\end{center}
\end{table*}

%%%%%%%%%%%%%%%%%%%%%%%% End TABLE E %%%%%%%%%%%%%%%%%%%%%%%%%%%%%%%%%
%%%%%%%%%%%%%%%%%%%%%%%%%%%%%%%%

%%%%%%%%%%%%%%%%%%%%%%%%%%%%%%%%
%%%%%%%%%%%%%%%%%%%%%%%% START Table: Word Embedding %%%%%%%%%%%%%%%%%%%%%%%%%%%
\begin{table*}[t]
\caption{Analysis of different distillation targets for speech reconstruction. (LM) indicates LM-guided distillation with averaged token representations, (CLS) indicates [CLS]-token-based distillation, (Baseline) refers to DM-Codec without any distillation, and (Word Embedding) indicates static BERT word embeddings. Contextual LM representations (LM, CLS) consistently outperform both the Baseline and static Word Embeddings. \textbf{Bold} highlights the best result and \underline{underline} the second-best result.}
\label{table:word-embedding}
\begin{center}
\small
\begin{tabular}{l cccc}
\toprule
\multicolumn{1}{c}{\bf Model} & \multicolumn{1}{c}{\bf WER $\downarrow$} & \multicolumn{1}{c}{\bf WIL $\downarrow$} & \multicolumn{1}{c}{\bf ViSQOL $\uparrow$} & \multicolumn{1}{c}{\bf STOI $\uparrow$} \\
\midrule
DM-Codec (Baseline)       & 4.97 & 8.02 & 2.95 & \textbf{0.935} \\
DM-Codec (LM)             & \textbf{4.36} & \textbf{7.06} & \textbf{3.18} & \textbf{0.935} \\
DM-Codec (CLS)            & \underline{4.47} & \underline{7.08} & \underline{3.12} & 0.926 \\
DM-Codec (Word Embedding) & 5.27 & 8.58 & 2.89 & 0.931 \\
\bottomrule
\end{tabular}
\end{center}
\end{table*}
%%%%%%%%%%%%%%%%%%%%%%%% END Table: Word Embedding %%%%%%%%%%%%%%%%%%%%%%%%%%%
%%%%%%%%%%%%%%%%%%%%%%%%%%%%%%%%

\subsection{\newtext{Ablation Study: Impact of Distillation Axis}}

\acltext{We experiment with different axes for feature distillation. In Eqn. \ref{eq:lm_distil} and \ref{eq:distil}, we calculate the distillation loss by maximizing cosine similarity along the feature dimension axis (D-axis). To evaluate the impact of an alternative approach, we conduct this ablation study by maximizing cosine similarity along the time dimension axis (T-axis). For this ablation, we modify Eqn. \ref{eq:lm_distil} and \ref{eq:distil} by computing cosine similarity across each time step instead of each feature dimension. Table \ref{table:axis} presents the complete results. We select D-axis {\pa} variants where the first Residual Vector Quantizer layer (RVQ-1) was used for LM distillation and all RVQ layers (RVQ-1:8) for SM distillation and retrain with the T-axis formulation.}

\acltext{\textbf{Results and Discussion:} In LM-guided distillation, the D-axis consistently achieves better performance across all metrics, with scores of 4.36 WER, 7.06 WIL, 3.18 ViSQOL, and 0.935 STOI, outperforming the relatively lower scores of the T-axis. Similarly, for combined LM and SM-guided distillation, the D-axis achieves superior results with 4.05 WER, 6.61 WIL, 3.26 ViSQOL, and 0.937 STOI compared to the T-axis.}
\acltext{These results demonstrate that calculating cosine similarity along the feature dimension axis (D-axis) is empirically more robust and effective. In contrast, cosine similarity along the time axis (T-axis) fails to capture contextual and semantic information. The identical WER of 4.76 for both LM and LM+SM distillation on the T-axis further highlights its limitations. However, our D-axis approach effectively incorporates richer information within each feature dimension, enhancing the model's capacity to capture nuanced patterns. This leads to improved speech reconstruction, validating our proposed method's robustness.}

\subsection{Ablation Study: Impact of Word Embedding Distillation Targets}
\label{sec:word-embeddings}

We conducted an experiment using static BERT word embeddings as the distillation target $L_{LM}$, referred to as DM-Codec (Word Embedding). Table~\ref{table:word-embedding} compares different distillation targets.

\textbf{Results and Discussion:} Both DM-Codec (LM) and DM-Codec (CLS), which distill contextual token representations from the LM, clearly outperform DM-Codec (Baseline) and DM-Codec (Word Embedding). In particular, DM-Codec (LM) achieves the best overall balance, with strong WER (4.36), WIL (7.06), ViSQOL (3.18), and STOI (0.935).  

By contrast, DM-Codec (Word Embedding), which relies on static embeddings, performs the worst across intelligibility metrics (WER 5.27, WIL 8.58) and shows limited improvement in perceptual quality (ViSQOL 2.89, STOI 0.931). This suggests that static word embeddings, lacking contextual representation, are insufficient as distillation targets for high-quality speech reconstruction. Overall, these results underscore the importance of contextualized LM representations in guiding effective distillation.
%%%%%%%%%%%%%%%%%%%%%%%%%%%%%%%%%%%%%%%

\section{DM-Codec Model Components}
\label{sec:components}
\textbf{Encoder Decoder.}
The encoder-decoder architecture in {\pa} is based on SEANet \citep{seanet}, leveraging the successful design employed in recent speech tokenization models \citep{speechtokenizer, encodec, soundstream}. The architecture is designed to efficiently process and reconstruct speech signals while maintaining high fidelity. The Encoder \( \mathbf{E} \) consists of a 1D convolution layer with \( \mathit{C} \)  channels and a kernel size of 7, followed by  \( \mathit{B} \) residual convolutional blocks. Each block contains a strided convolutional downsampling layer with kernel size \( \mathit{K} \) (where \( \mathit{K}=2\mathit{S} \) , and \( \mathit{S} \) represents the stride), paired with a residual unit. The residual unit comprises two convolutional layers with a kernel size of 3 and a skip connection, while the number of channels is doubled at each downsampling stage.  This is followed by a two-layer BiLSTM and a final 1D convolutional layer with \( \mathit{D} \) output channels and a kernel size of 7.  The Decoder \( \mathbf{D} \) mirrors the encoder’s structure but replaces BiLSTM with LSTM, strided convolutions with transposed convolutions, and employs reversed strides for up-sampling. The final audio output is reconstructed from \( \mathbf{D} \). For the experiments, we use the following configuration: \( \mathit{C} \)  = 32, \( \mathit{B} \) = 4, and \( \mathit{S} \) = (2, 4, 5, 8).

\textbf{Residual Vector Quantizers.}
The Residual Vector Quantizer (RVQ) plays a central role in our tokenization process, quantizing the encoder's outputs. Our implementation is inspired by the training procedures described in Encodec \citep{encodec} and SpeechTokenizer \citep{speechtokenizer}. The RVQ projects input vectors to the most similar entry in a codebook, and	the residual is calculated and processed in subsequent quantization steps, each utilizing a different codebook. The codebook entries are updated using an \textit{exponential moving average} (EMA) with a \textit{decay rate} of 0.99 for the matched item, while unmatched entries are replaced by candidates from the current batch.  To ensure proper gradient flow during training, we employ a \textit{straight-through estimator}. A \textit{commitment loss} is also computed and added to the total training loss to promote stability.

\textbf{Discriminators.}
We incorporate a trio of discriminators to enhance the quality and realism of the generated speech: the Multi-Scale Discriminator (MSD), the Multi-Period Discriminator (MPD), and the Multi-Scale Short-Time Fourier Transform (MS-STFT) discriminator. The MS-STFT discriminator follows the implementation outlined in \citep{encodec}, operating on the real and imaginary components of multi-scale complex-valued STFTs. It begins with a 2D convolutional layer, followed by 2D convolutions with increasing dilation rates in the time dimension (1, 2, and 4) and a stride of 2 across the frequency axis in each sub-network. A final 2D convolution with a kernel size of 3 × 3 and a stride of (1, 1) is applied to produce the prediction. The MSD and MPD discriminators follow the architectures introduced in \citep{hifigan}, with adjustments to the channel numbers to align the parameter count more closely with the MS-STFT discriminator. This ensemble of discriminators works in concert to provide comprehensive feedback on various aspects of the generated speech, contributing to the overall quality and naturalness of the output.

\section{\newtext{Human Evaluation Methodology}}
\label{sec:human-eval}
\newtext{To evaluate the quality and effectiveness of our approach, we conducted human evaluations using Mean Opinion Score (MOS) and Similarity Mean Opinion Score (SMOS) metrics, following methodologies established in prior works such as SpeechTokenizer and Vall-E. The study was conducted under an approved Institutional Review Board (IRB) protocol to ensure ethical compliance and participant safety. A total of 50 proficient English speakers, comprising graduate and undergraduate students, were selected as evaluators based on their high language comprehensibility. These participants volunteered for the evaluation, were briefed on the study’s purpose, and were provided no information that could bias their judgments. }

\newtext{The evaluation process involved each participant rating batches of fully anonymized and randomized speech samples via a web-based survey interface, with clear and standardized guidelines to ensure consistent and unbiased scoring. Each batch contained 16 samples, including outputs from both our proposed models and baseline systems. For the speech reconstruction task, participants rated the perceptual quality of the speech samples using the MOS, based on criteria such as naturalness, intelligibility, and clarity, employing a 5-point Likert scale, where higher scores indicated superior quality. For the text-to-speech evaluation, participants provided two distinct ratings: MOS, to measure the overall naturalness of the generated speech, and SMOS, to evaluate the similarity of the generated speech to the original speaker’s voice, both rated on a 1-to-5 scale with 1-point increments. To enhance reliability and mitigate individual evaluator bias, each sample was rated by multiple participants. Figure \ref{fig:survey_interface} shows the interface view and instructions given to participants.}

%%%%%%%%%%%%%%%%%%%%%%%%%%%%%%%%%%%%%%%%%%%%%%%%%%%%%%%%%%%%
% PAGE 1: Views 1 and 2
\begin{figure*}[t]
    \centering
    \begin{subfigure}[b]{0.7\textwidth}
        \centering
        \includegraphics[width=\textwidth]{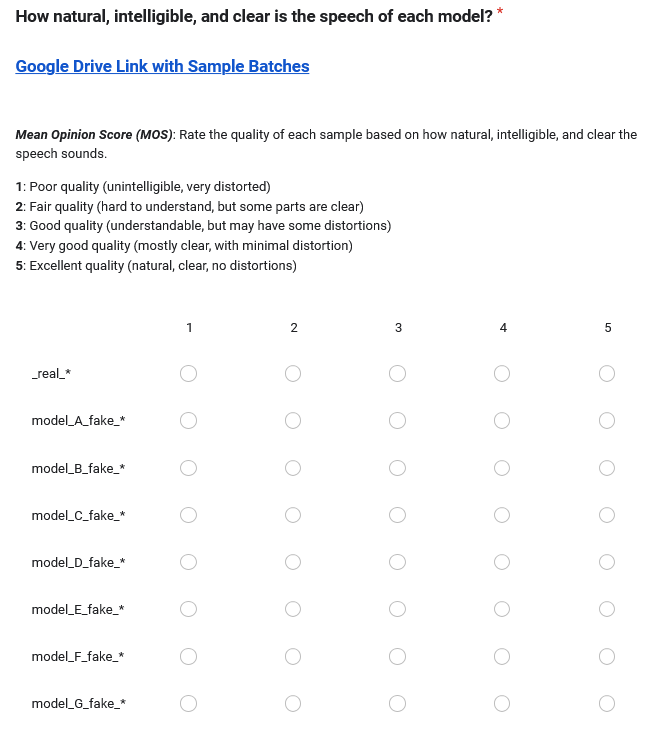}
        \caption*{Interface View 1}
        \label{fig:subfig1}
    \end{subfigure}

    % \vspace{0.5cm}

    \begin{subfigure}[b]{0.5\textwidth}
        \centering
        \includegraphics[width=\textwidth]{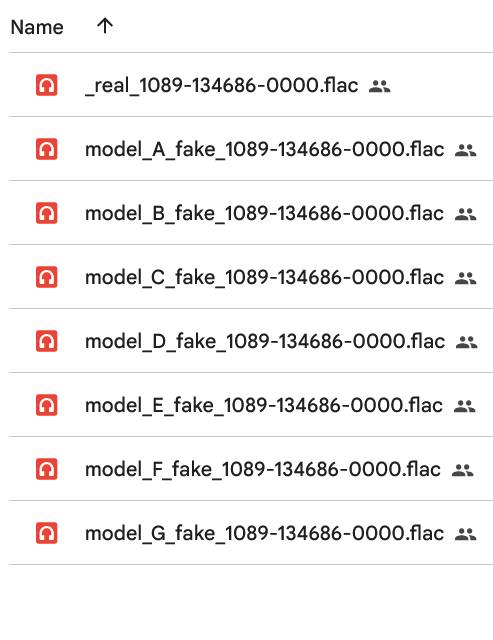}
        \caption*{Interface View 2}
        \label{fig:subfig2}
    \end{subfigure}
\end{figure*}

% PAGE 2: Views 3, 4, and 5 (caption at the end)
\begin{figure*}[t]\ContinuedFloat
    \centering
    \begin{subfigure}[b]{0.48\textwidth}
        \centering
        \includegraphics[width=\textwidth]{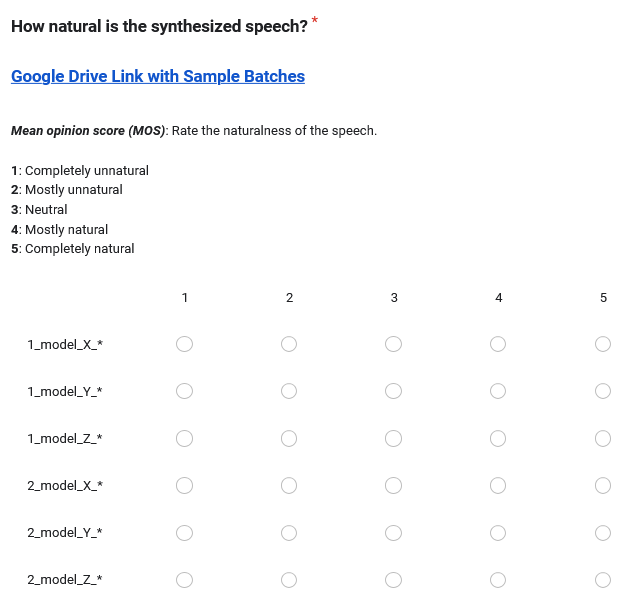}
        \caption*{Interface View 3}
        \label{fig:subfig3}
    \end{subfigure}
    \hfill
    \begin{subfigure}[b]{0.48\textwidth}
        \centering
        \includegraphics[width=\textwidth]{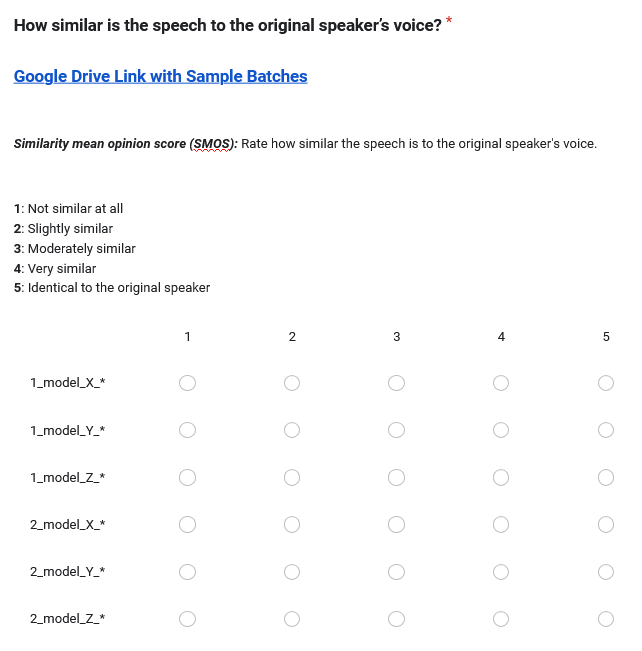}
        \caption*{Interface View 4}
        \label{fig:subfig4}
    \end{subfigure}

    % \vspace{0.5cm}

    \begin{subfigure}[b]{0.55\textwidth}
        \centering
        \includegraphics[width=\textwidth]{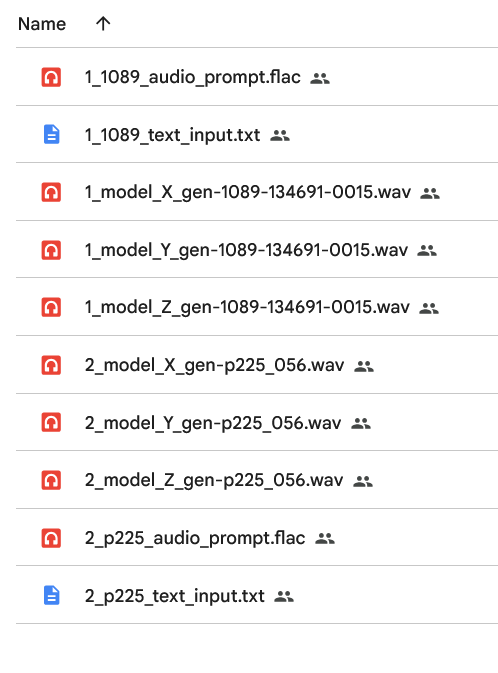}
        \caption*{Interface View 5}
        \label{fig:subfig5}
    \end{subfigure}

    \caption{Web-based survey interface and questionnaire used for human evaluation of samples.}
    \label{fig:survey_interface}
\end{figure*}
%%%%%%%%%%%%%%%%%%%%%%%%%%%%%%%%%%%%%%%%%%%%%%%%%%%%%%%%%%%%

\section{Broader Impact and Potential Risks}
% \noindent \textbf{Broader Impact.} 
The integration of language models in speech processing has traditionally focused on model-specific implementations or specific training objectives. In this work, we propose a novel approach by leveraging language models during the tokenization phase through our model, DM-Codec. By incorporating language-specific representations from the corresponding text, DM-Codec enhances the quality of discrete speech representations. This method bridges the gap between language and speech models, offering a more unified approach to multimodal representation learning. DM-Codec provides a robust framework for generating high-quality audio representations, with potential applications in various domains, including multilingual speech processing, low-resource languages, and other audio-related tasks. Our findings pave the way for more effective and contextually aware speech processing models, contributing to advancements in the broader field of speech and language technologies. Despite its advancements, DM-Codec-TTS poses risks of misuse, such as voice spoofing or impersonation, due to its ability to synthesize speech while preserving speaker identity. Our future work will prioritize robust detection mechanisms and ethical safeguards to ensure responsible use.

\section{Reconstructed Speech Comparison}

\begin{figure*}[ht] % Use * to span two columns
    \centering
    % First row of figures
    \begin{subfigure}[b]{0.425\textwidth}
        \centering
        \href{https://drive.google.com/file/d/1QdLH15Sx-F__C3NjsDe_7dPeg6U5JuEh/view?usp=drive_link}{%
            \includegraphics[width=0.1\textwidth]{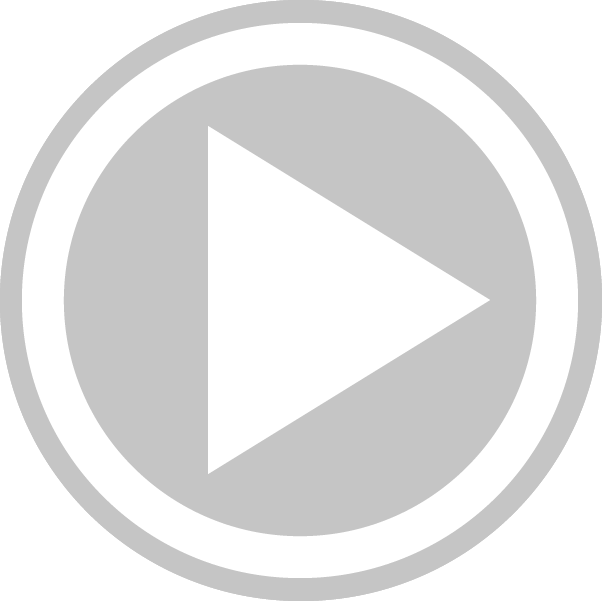}}
        \includegraphics[width=\textwidth]{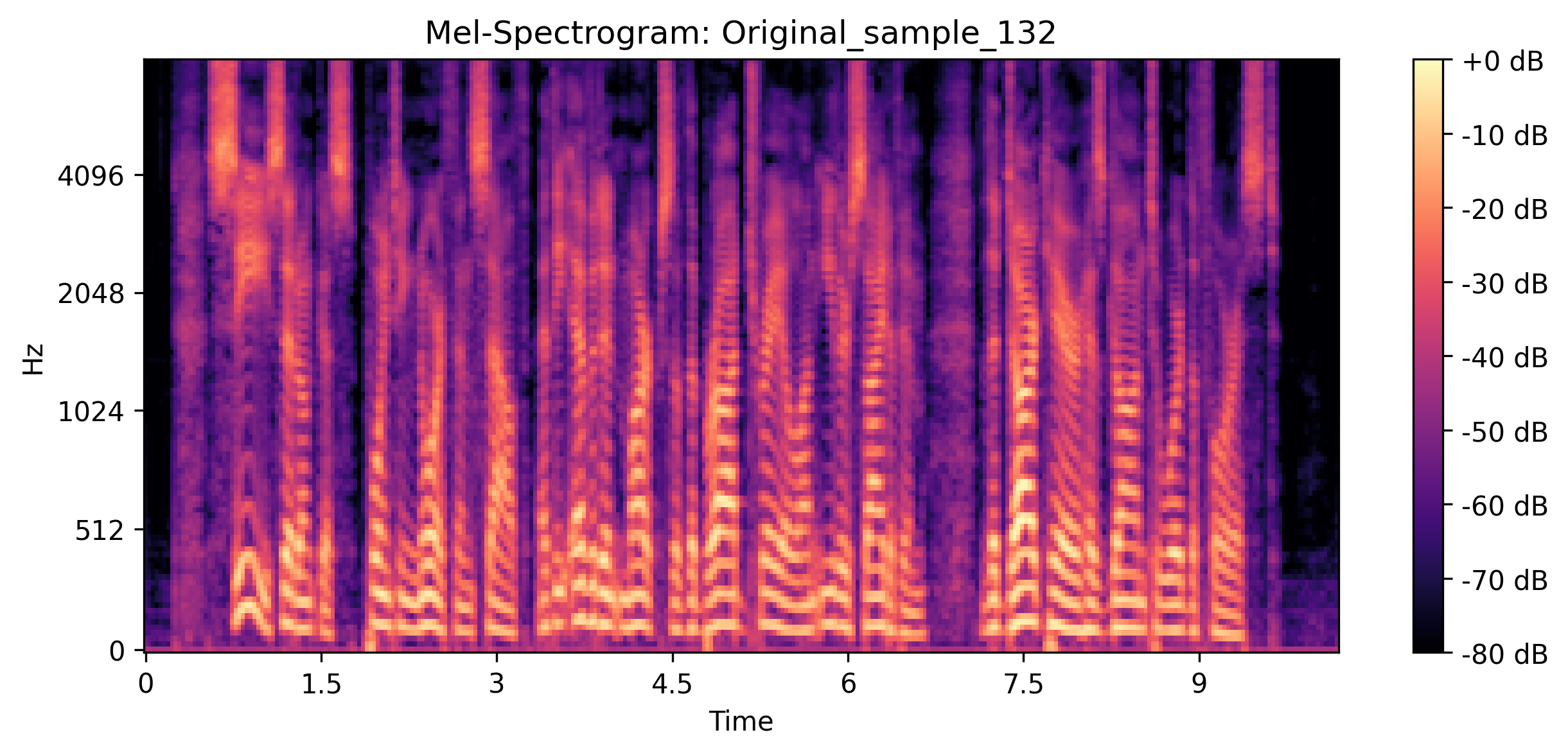}
        \caption{Original Speech 1\label{fig:original1}}
    \end{subfigure}
    \hfill
    \begin{subfigure}[b]{0.425\textwidth}
        \centering
        \href{https://drive.google.com/file/d/13hFzjeMPog99rcYbxroCwSJEA3i4TReU/view?usp=drive_link}{%
            \includegraphics[width=0.1\textwidth]{figures/play_button.png}}
        \includegraphics[width=\textwidth]{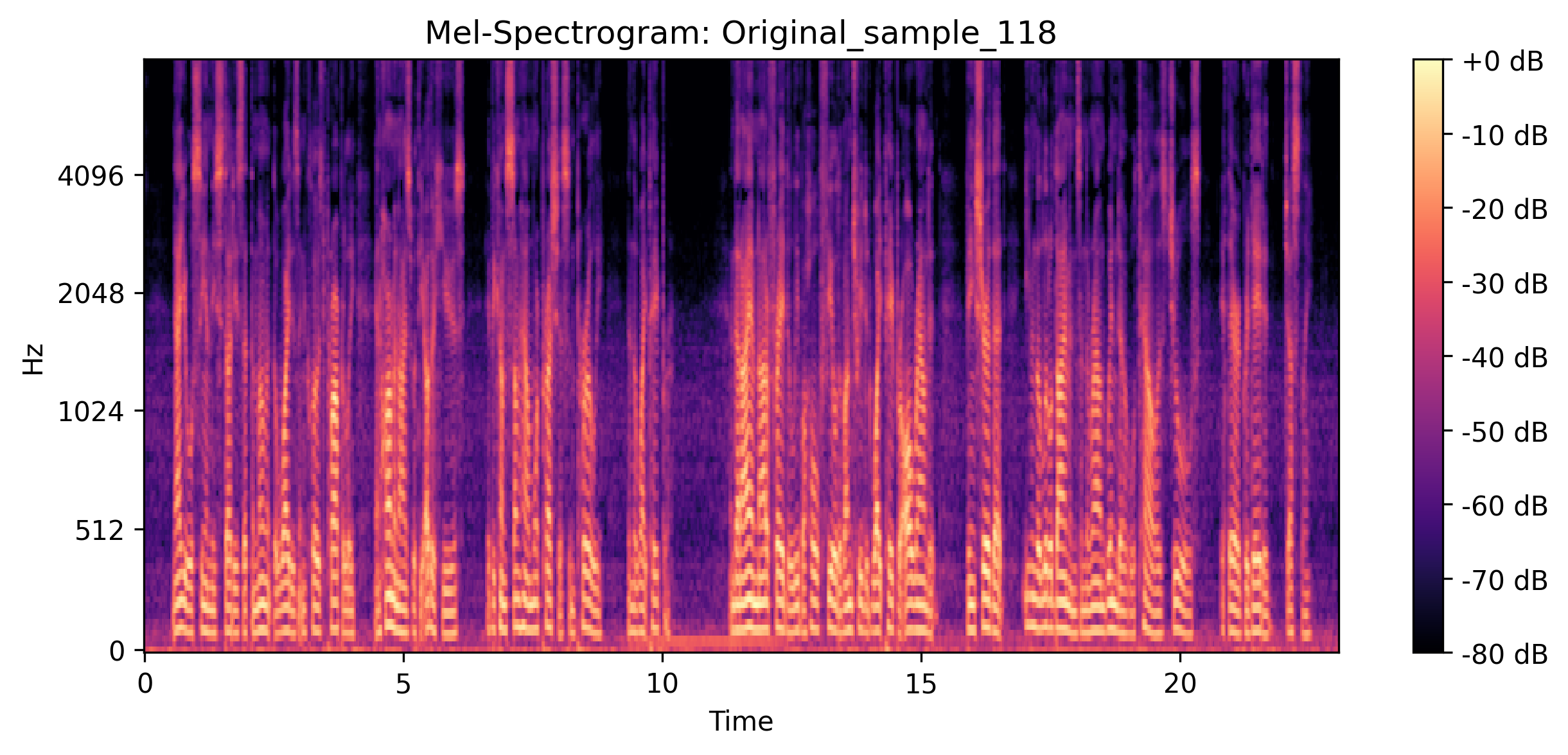}
        \caption{Original Speech 2\label{fig:original2}}
    \end{subfigure}

    % Second row of figures
    % \vspace{10pt}
    \begin{subfigure}[b]{0.425\textwidth}
        \centering
        \href{https://drive.google.com/file/d/1RW3acjA4nel75u3zpGGL59sS4xb6kp1q/view?usp=drive_link}{%
            \includegraphics[width=0.1\textwidth]{figures/play_button.png}}
        \includegraphics[width=\textwidth]{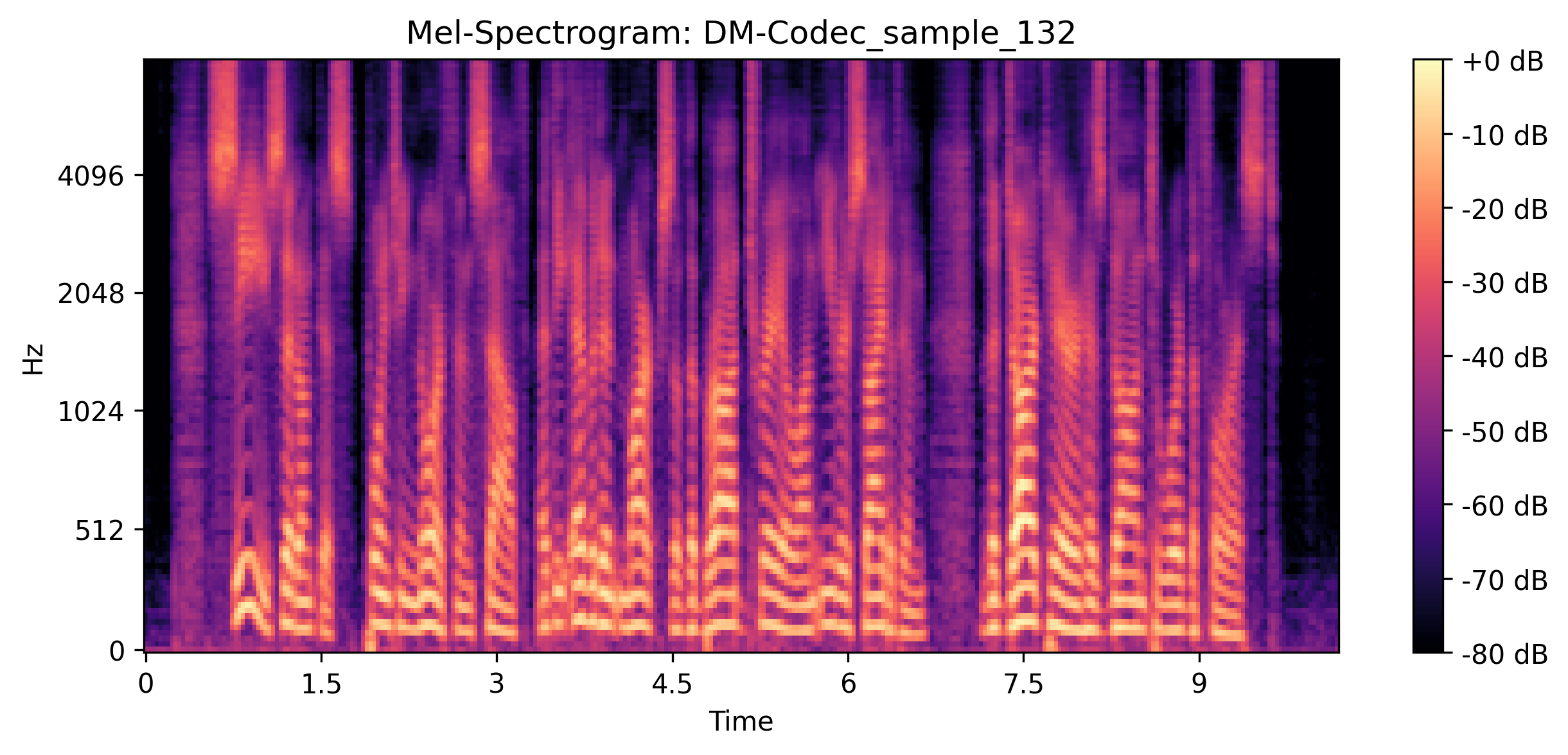}
        \caption{DM-Codec Speech 1\label{fig:dm_codec1}}
    \end{subfigure}
    \hfill
    \begin{subfigure}[b]{0.425\textwidth}
        \centering
        \href{https://drive.google.com/file/d/1Zv58qXHB4UZDcOE9MMfn0w5ve7Hw2wyY/view?usp=drive_link}{%
            \includegraphics[width=0.1\textwidth]{figures/play_button.png}}
        \includegraphics[width=\textwidth]{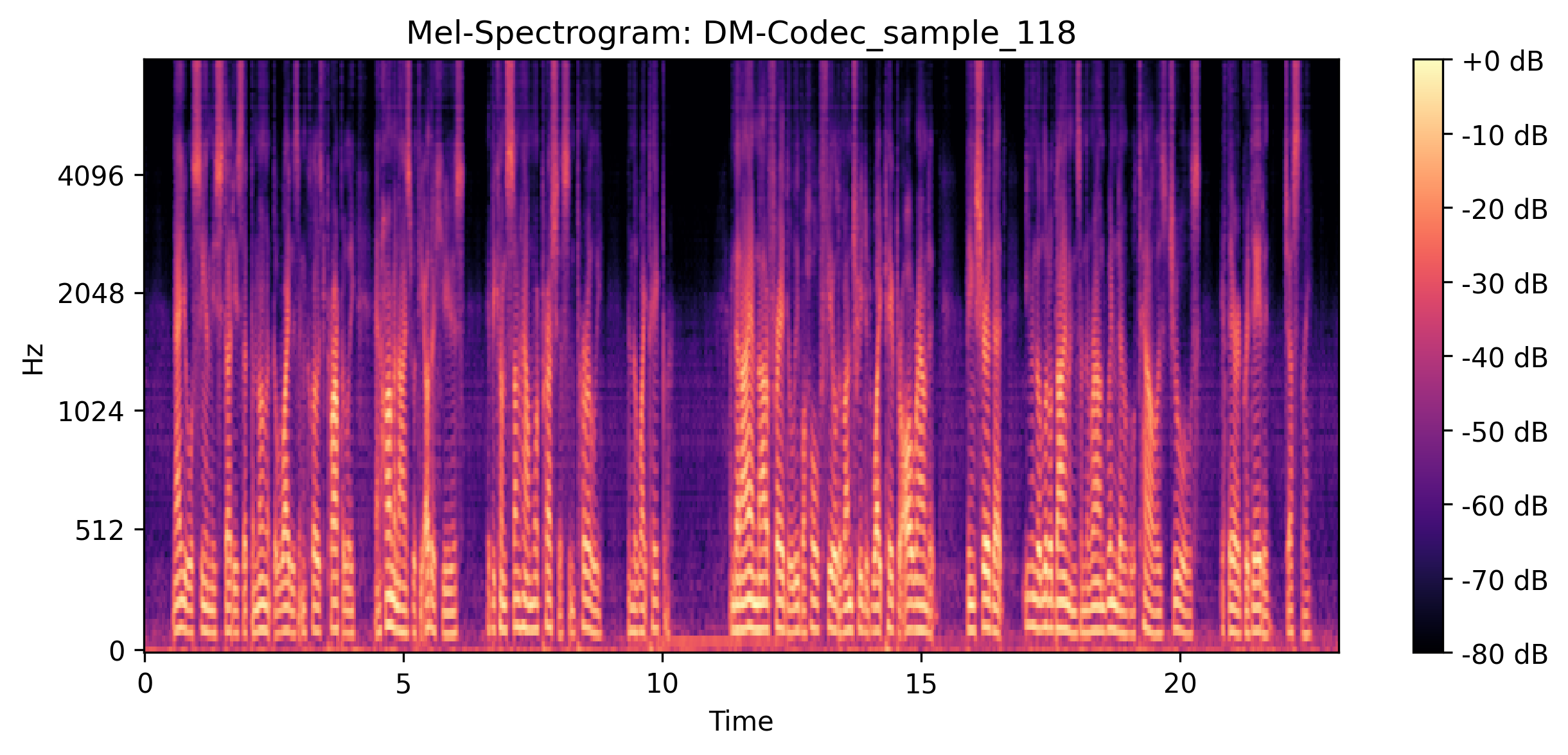}
        \caption{DM-Codec Speech 2\label{fig:dm_codec2}}
    \end{subfigure}

    % Third row of figures
    % \vspace{10pt}
    \begin{subfigure}[b]{0.425\textwidth}
        \centering
        \href{https://drive.google.com/file/d/1005eTwIVDa4zmCx_5v3JVGpKK-Zh91xF/view?usp=drive_link}{%
            \includegraphics[width=0.1\textwidth]{figures/play_button.png}}
        \includegraphics[width=\textwidth]{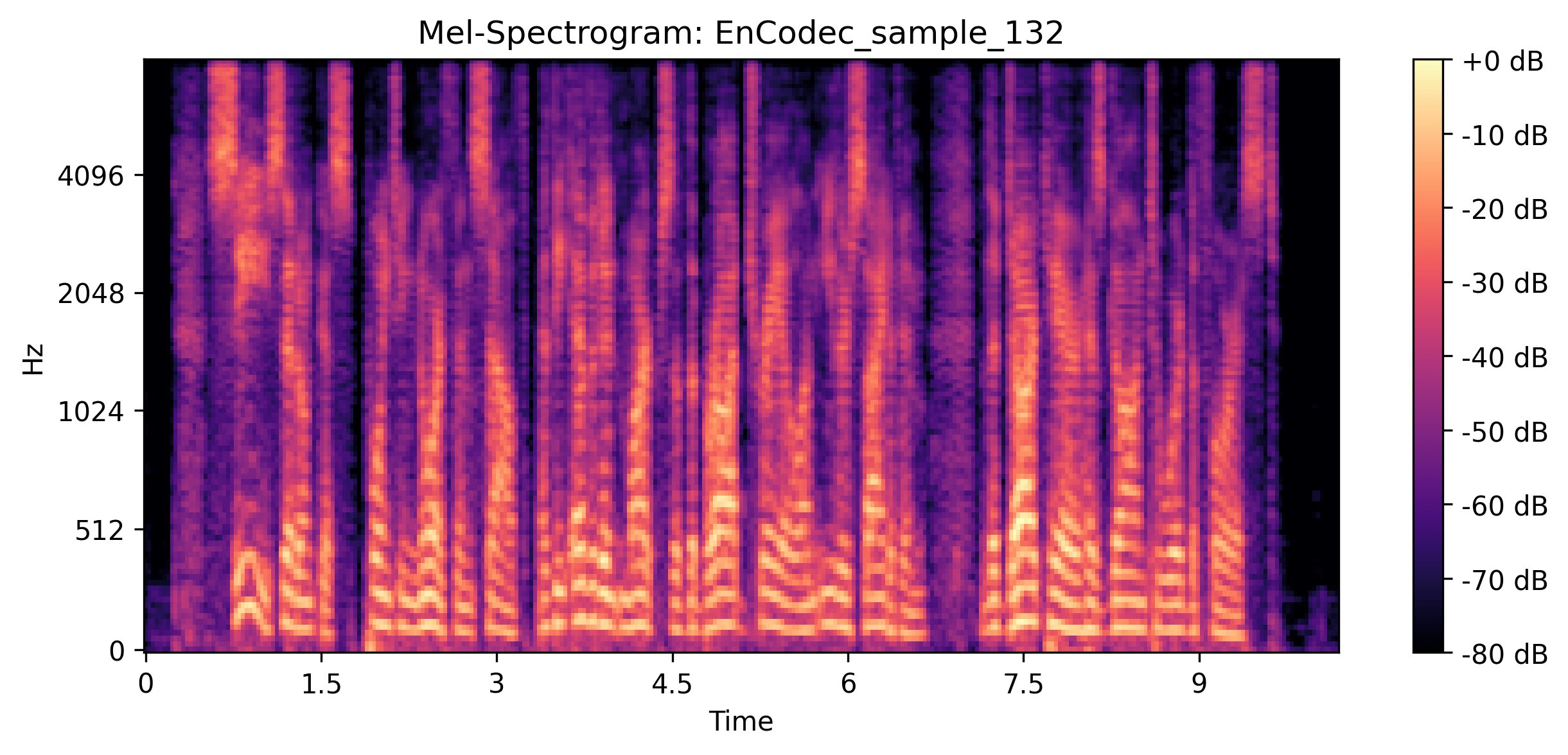}
        \caption{EnCodec Speech 1\label{fig:encodec1}}
    \end{subfigure}
    \hfill
    \begin{subfigure}[b]{0.425\textwidth}
        \centering
        \href{https://drive.google.com/file/d/1OV_WbvkamG2JRNOjpRQYlPFiae_nNpcf/view?usp=drive_link}{%
            \includegraphics[width=0.1\textwidth]{figures/play_button.png}}
        \includegraphics[width=\textwidth]{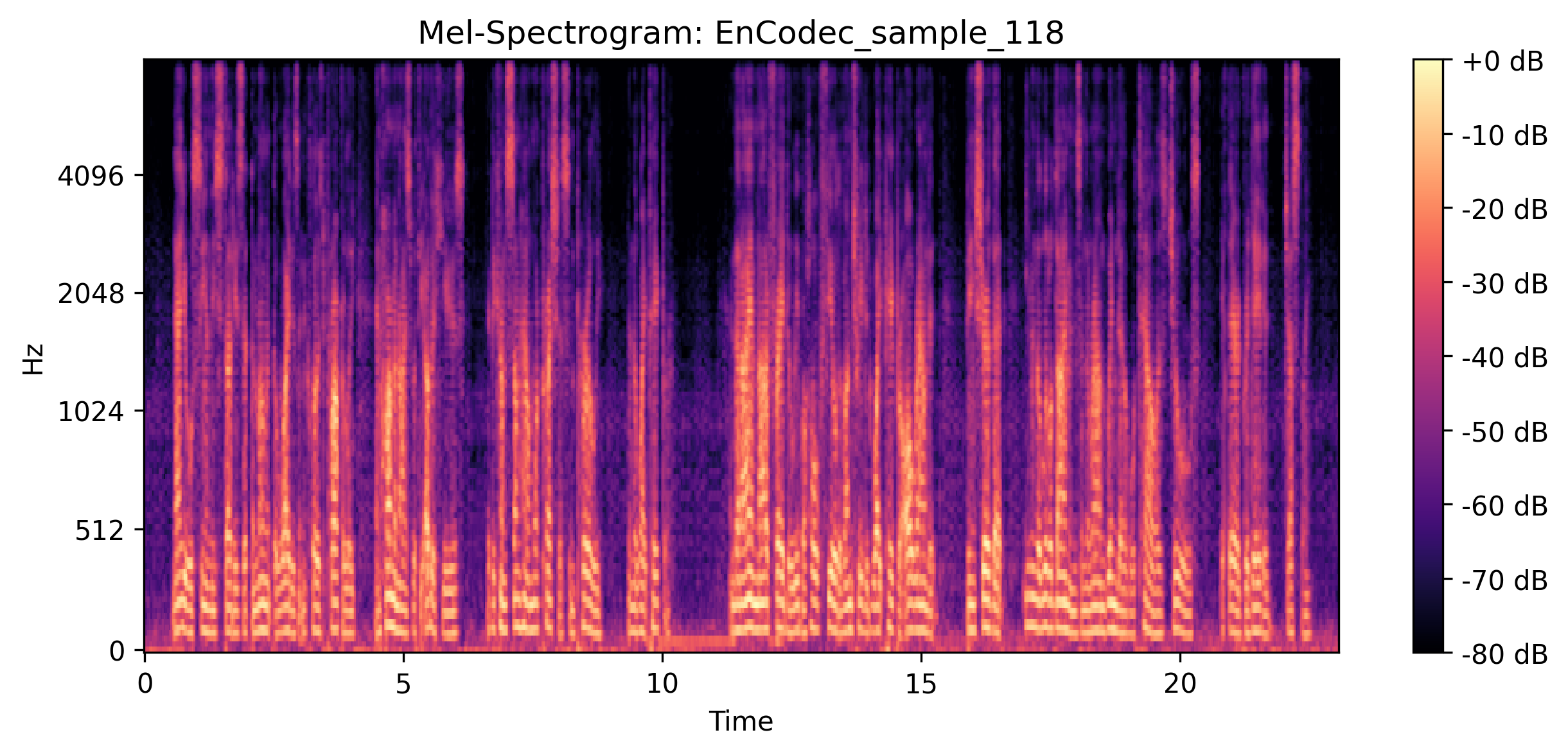}
        \caption{EnCodec Speech 2\label{fig:encodec2}}
    \end{subfigure}

    % Fourth row of figures
    % \vspace{10pt}
    \begin{subfigure}[b]{0.425\textwidth}
        \centering
        \href{https://drive.google.com/file/d/10ngPpXvc_cZIu-7GTB4ZFDFBr9A1cRDA/view?usp=drive_link}{%
            \includegraphics[width=0.1\textwidth]{figures/play_button.png}}
        \includegraphics[width=\textwidth]{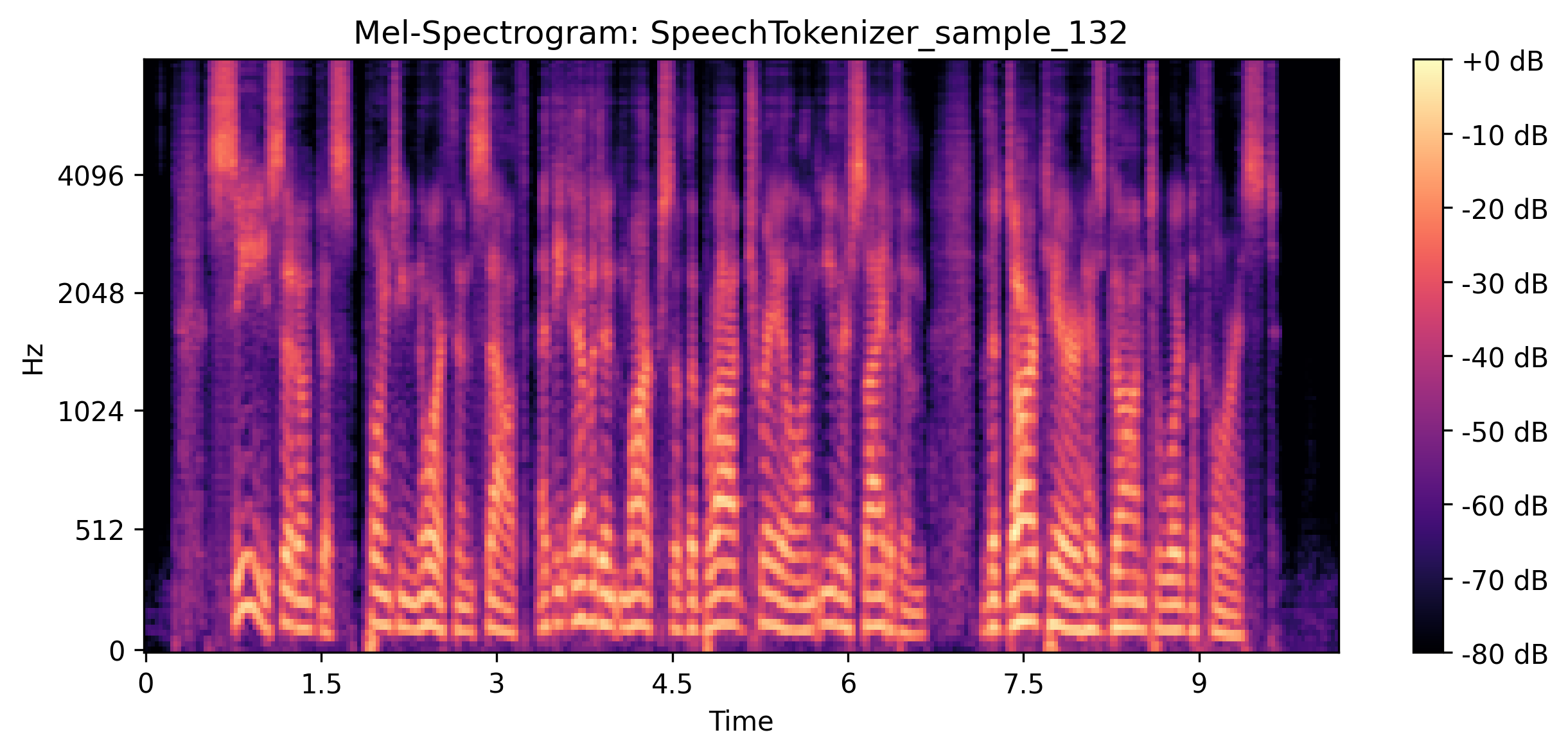}
        \caption{SpeechTokenizer Speech 1\label{fig:speech_tokenizer1}}
    \end{subfigure}
    \hfill
    \begin{subfigure}[b]{0.425\textwidth}
        \centering
        \href{https://drive.google.com/file/d/1oVOghmQNeGFo2oL09tw4nt2LVIApiXyp/view?usp=drive_link}{%
            \includegraphics[width=0.1\textwidth]{figures/play_button.png}}
        \includegraphics[width=\textwidth]{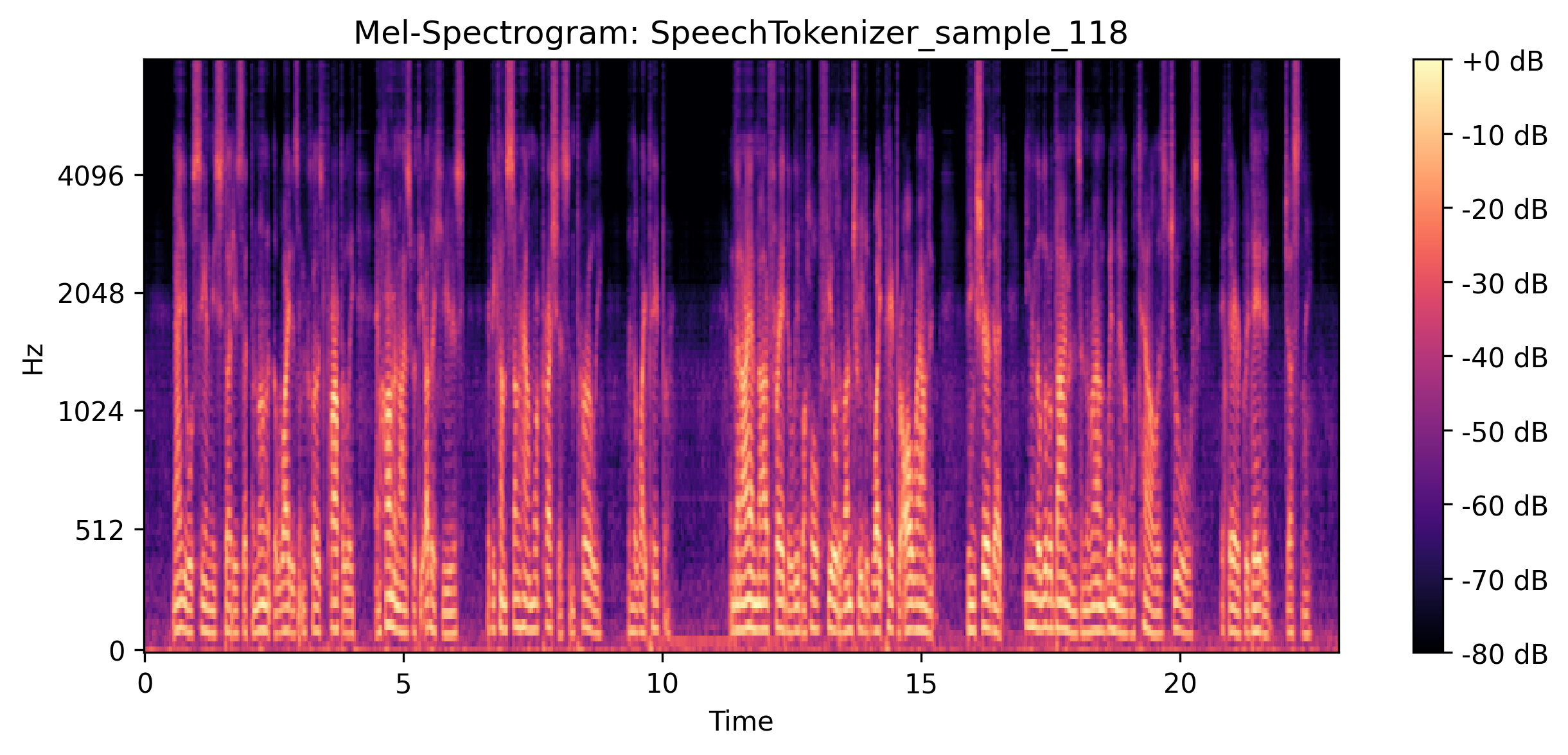}
        \caption{SpeechTokenizer Speech 2\label{fig:speech_tokenizer2}}
    \end{subfigure}

    % Fifth row of figures
    % \vspace{10pt}
    \begin{subfigure}[b]{0.425\textwidth}
        \centering
        \href{https://drive.google.com/file/d/1D7dOgR2FccCPU33vS7UCp5XhiEFgGWIa/view?usp=drive_link}{%
            \includegraphics[width=0.1\textwidth]{figures/play_button.png}}
        \includegraphics[width=\textwidth]{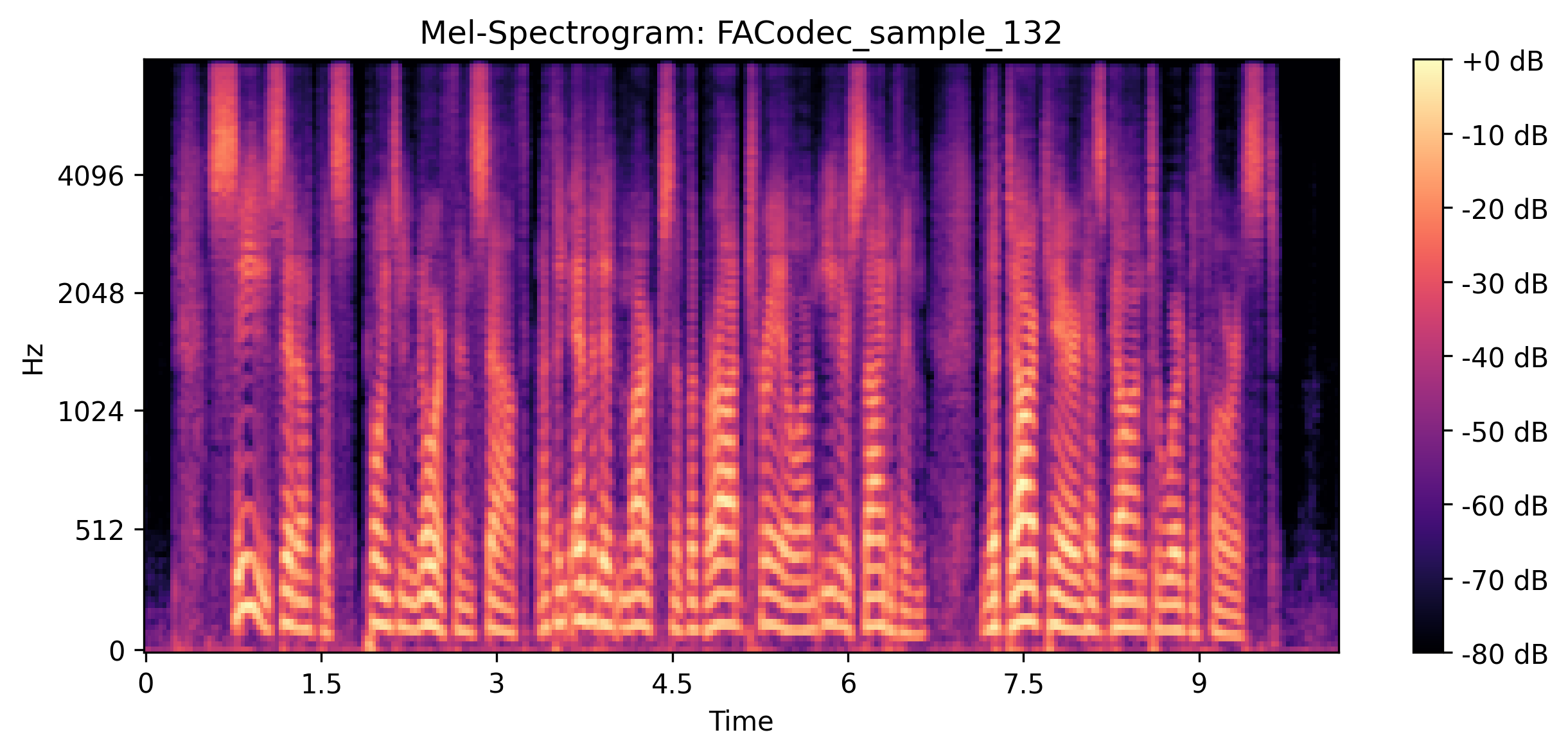}
        \caption{FACodec Speech 1\label{fig:fa_codec1}}
    \end{subfigure}
    \hfill
    \begin{subfigure}[b]{0.425\textwidth}
        \centering
        \href{https://drive.google.com/file/d/1CCvqCU_A3eIAkC8TGWIi64kCNAQ4hIQl/view?usp=drive_link}{%
            \includegraphics[width=0.1\textwidth]{figures/play_button.png}}
        \includegraphics[width=\textwidth]{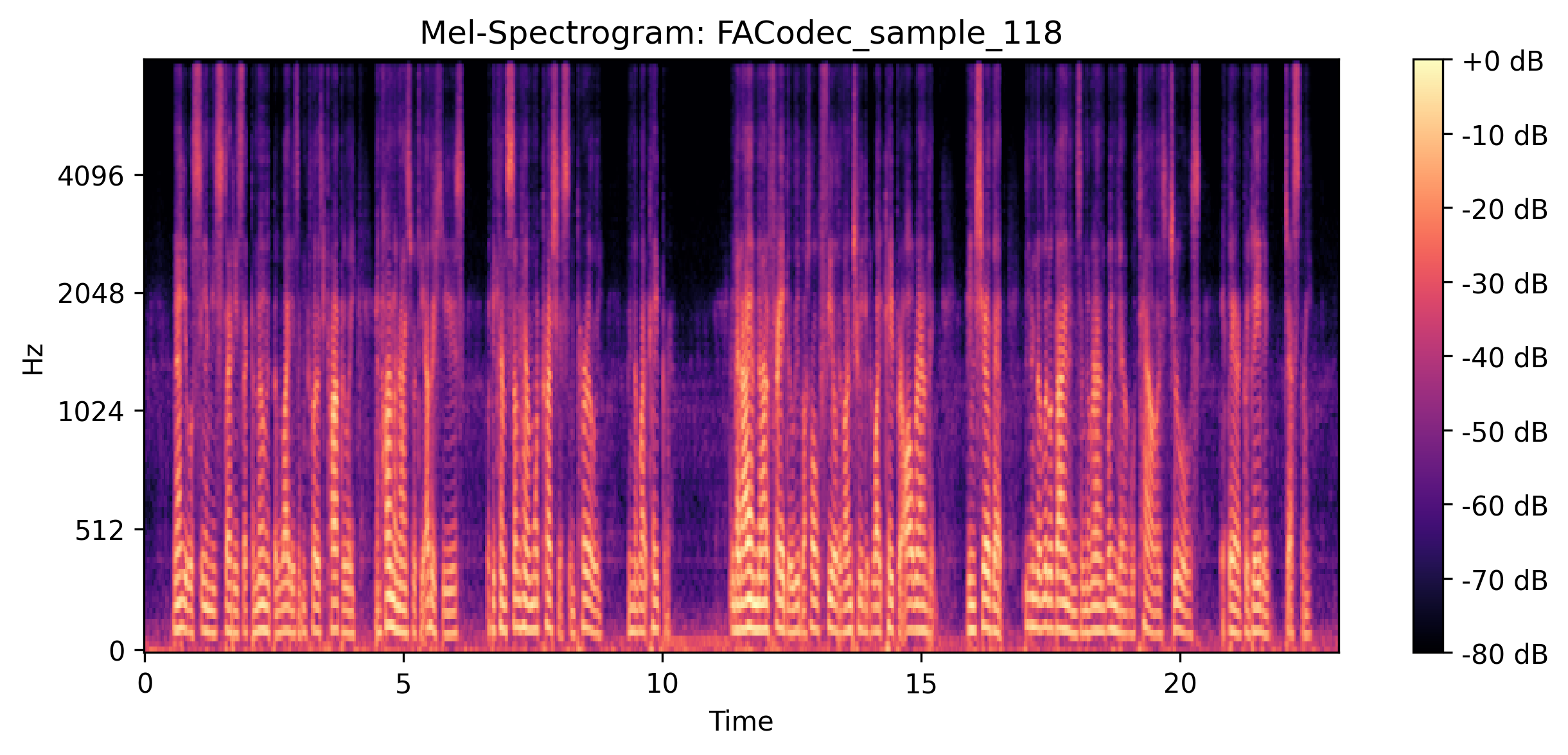}
        \caption{FACodec Speech 2\label{fig:fa_codec2}}
    \end{subfigure}

    \caption{Reconstructed speech examples with clickable play buttons above each Mel-spectrogram.}
    \label{fig:images_layout}
\end{figure*}

Figure \ref{fig:images_layout} compares the Mel-Spectrograms of the original speech with reconstructed speech from {\pa}, EnCodec, SpeechTokenizer, and FACodec. 

\end{document}